\PassOptionsToPackage{table}{xcolor}
\documentclass[11pt]{article}
\usepackage[T1]{fontenc}

\usepackage{amsmath}
\usepackage{amssymb}
\usepackage{bm}
\usepackage[cal=boondox]{mathalfa}

\usepackage{cite}
\usepackage{url}
\usepackage{hyperref}

\usepackage{graphicx}
\usepackage{booktabs}
\usepackage{tabularx}
\usepackage{array}
\usepackage{colortbl}
\usepackage{multirow}
\usepackage{makecell}
\usepackage{rotating}
\usepackage[font=small,labelfont=bf,tableposition=top]{caption}
\usepackage{subcaption}

\usepackage{algorithm}
\usepackage{algpseudocode}

\usepackage{authblk}
\usepackage{xfrac}
\usepackage{color}
\usepackage{soul}
\usepackage{xcolor}
\usepackage{enumitem}
\usepackage[section]{placeins}
\usepackage{lineno}
\usepackage{siunitx}
\usepackage{etoolbox}
\usepackage{comment}
\usepackage[text={16cm,23cm},centering]{geometry}

\DeclareCaptionFormat{subfig}{\figurename~#1#2#3}
\DeclareCaptionSubType*{figure}
\DeclareCaptionType{equ}[List of Equations][]

\newcolumntype{P}[1]{>{\centering\arraybackslash}p{#1}}

\newcommand{\ve}[1]{\pmb{#1}}

\newcommand{\best}[1]{\mathbf{#1}}
\newcommand{\forceemph}[1]{{\itshape #1}}
\newcommand{\set}[1]{\mathcal{#1}}
\newcommand{\class}{\mathrm{class}}
\newcommand{\sign}{\mathrm{sign}}

\DeclareMathOperator{\trace}{tr}
\DeclareMathOperator{\diag}{diag}

\newcommand{\KA}{\mathrm{KA}}
\newcommand{\CKA}{\mathrm{CKA}}
\newcommand{\LKA}{\mathrm{LKA}}
\newcommand{\HKA}{\mathrm{HKA}}

\newcommand{\HSIC}{\mathrm{HSIC}}

\newcommand{\RFSC}{\mathrm{RFSC}}
\newcommand{\GL}{\mathrm{GL}}
\newcommand{\CS}{\mathrm{CS}}

\newcommand{\Kcent}{\ve{K}_{\mathrm{cent}}}
\newcommand{\Kmod}{\ve{H}}
\newcommand{\Kminmod}{\ve{H}_{\mathrm{min}}}
\newcommand{\Kideal}{\ve{K}_{\mathrm{ideal}}}
\newcommand{\Kident}{\ve{K}_{\mathrm{ident}}}
\newcommand{\Kmin}{\ve{K}_{\mathrm{min}}}
\newcommand{\Knorm}{\ve{K}_{\mathrm{norm}}}

\newcommand{\hilspace}{\mathcal{H}}

\newcommand{\frobprod}[2]
    {\left\langle\!\left\langle #1,#2
    \right\rangle\!\right\rangle}

\newtheorem{definition}{Definition}

\def\BibTeX{%
    {\rm B\kern-.05em{\sc i\kern-.025em b}\kern-.08em
    T\kern-.1667em\lower.7ex\hbox{E}\kern-.125emX}%
}

\begin{document}
\title{The Confidence Score: a New Measure on Kernels}

\author[1]{Elizabeth Hofer}
\author[2]{Martin v. Mohrenschildt}

\affil[1]{Department of Computing and Software, McMaster University, Hamilton, Ontario, Canada (e-mail: hofere1@mcmaster.ca)}
\affil[2]{Department of Computing and Software, McMaster University, Hamilton, Ontario, Canada (e-mail: mohrens@mcmaster.ca)}

\date{}

\maketitle

\begin{abstract}
Across a wide range of applications, kernel measures are used to tune hyperparameters, select among candidate kernels, and assess the suitability of kernels for specific learning tasks. 
For classification, kernel measures are used to predict which kernels may produce accurate kernel classifiers.
Many existing kernel measures, such as alignment to the ideal kernel, are designed to reward kernels which captures all samples from the same class as similar.
While effective in many settings, this requirement is unnecessarily restrictive for classification, where classes may exhibit substantial internal structure.
%Within the context of classification, alignment is usually measured between a candidate kernel and the ideal kernel  --- the kernel that assigns equal similarity to all pairs of samples from the same class and equal dissimilarity to all pairs of samples from different classes.
%While measures that capture the alignment to the ideal kernel have proven effective in many applications, they tend to assign higher scores to kernels that collapse within-class structure, and may therefore penalize kernels that preserve meaningful intra-class variability.
In this work, the `Confidence Score', a new kernel measure, is proposed. 
Unlike alignment-to-the-ideal kernel measures, the Confidence Score measures the alignment between an encoding of the ground truth and that ground truth projected by the kernel, allowing kernels that preserve intra-class structure to achieve high scores. 
%\hl{The Confidence Score is parameter-free, reflecting the principle that kernel measures should not require additional tuning.}
%\hl{In contrast to some variants of alignment-to-the-ideal kernel measures designed for a similar purpose, the Confidence Score does not depend on additional tunable parameters.}
Through experiments, it is demonstrated that the Confidence Score exhibits a higher correlation with classification accuracy across a variety of kernels on both real-world and synthetic datasets, compared to select other kernel measures.
\end{abstract}

\section{Introduction}
\label{sec:introduction}
Similarity is a fundamental concept in machine learning: similar things are clustered together, and similarity is the key concept in classification. %martins sentence
The kernel, representing the inner product in some Hilbert space, is a mathematical tool that is used successfully in this domain to capture similarity between points \cite{zhang2007local, gillespie2001optimizing, cortes2010two, hofmann2008kernel, du2024exploring, wang2017multiple, kumar2012binary,zhong2013optimizing}. %martins sentence
Learning algorithms obtain their information from the similarity encoded by the kernel; thus the potential performance can be assessed from a measure on that kernel. %martins sentence

Many approaches of generating kernels are possible; entire, potentially infinite families of kernels can be constructed. %matrin sentence
For example, many kernels have parameters that can be tuned (e.g. bandwidth)\cite{keerthi2003asymptotic, camargo2009multi,zhong2013optimizing, tian2015efficient}, and some kernels are created by selecting and combining kernels into a new kernel, this is known as Multiple Kernel Learning (MKL) \cite{cristianini2001kernel, kandola2002optimizing, lanckriet2004learning, kumar2012binary, lu2014multiple, wang2017multiple, wang2018two, kwok2003learning, cortes2009learning}. %for example the bandwidth of a kernel significantly affects its classification performance. 
This creates an infinitely large search space of possible kernels. %for any ML task. 
While kernels themselves \forceemph{measure} similarity between data points, a \forceemph{measure on kernels} is needed to evaluate a candidate kernel pulled from this search space. 
%There are myriad ways to construct a kernel, Candidate kernels (pulled from {\it families of kernels}) can be constructed in several different ways, 

For classification, the objective of a kernel measure is to identify kernels or kernel parameters that maximize accuracy. 
The evaluation of the kernel measure should be computationally less expensive than computing the classification accuracy on a training dataset.
%This work presents an efficient-to-compute kernel measure that will predict how well a kernel will potentially perform in a classification task without running the expensive learning algorithm. %martins sentence
Kernel measures and their applications, ranging from classification to neural networks, have been extensively studied in the literature. 

Kernel Alignment \cite{cristianini2001kernel}, and its notable refinement Centred Kernel Alignment (CKA) \cite{cortes2009learning}, measures how well a kernel `aligns' with a learning target.
The learning target is usually the {\it ideal kernel}, the kernel that measures the similarity between two points of the same class as $1$, and $-1$ otherwise.
%Using CKA to optimize parameters of a family of kernels often does not necessarily produce the most accurate kernel classifiers.
%This is particularly true when the class-conditional distribution is multimodal.
To measure alignment with the ideal kernel is to assume that all points belonging to the same class are similar; that is, `similar $\leftrightarrow$ same class'.
It is known that the CKA is not well-correlated with classification accuracy \cite{wang2017multiple}. %matrins sentence
This problem is particularly pronounced in cases where the data is distributed following a multimodal class distribution, for example, if data belonging to one class is distributed across many isolated clusters, because the assumption of `similar $\leftrightarrow$ same class' does not hold in this case.
The kernel measure proposed in this work follows the `weaker' assumption: if data points are similar, they should belong to the same class; that is, `similar $\rightarrow$ same class'.
%We formulate as this: the ideal kernel assumes that all data points from the same class should be similar, but we work on the assumption that {\it data points that are similar should be from the same class}.  
%So a kernel, seen as a correlation, is good if for each data point correlation to elements of the same class is stronger than the correlation to data points from different classes. 
%This principle aligns with the intuition underlying weighted neighbourhood-based classifiers such as KNN.
%Under this principle, which is the key idea in weighted neighbourhood based classification algorithms such as KNN work.
%From this perspective, a kernel is considered effective if, for each data point, similarity to points of the same class exceeds similarity to points of different classes.

We demonstrate that a kernel measure based on the `weaker' assumption allows for the identification of more suitable kernels for classification.
After a review of the mathematics behind kernels in Section~\ref{sec_background_math}, the literature review follows in Section~\ref{sec_related_work_measures}.
In Section~\ref{sec_a_new_measure}, the \forceemph{Confidence Score} --- a new kernel measure --- is introduced, and Section~\ref{sec_eval} presents experimental results evaluating its performance. 
The discussion of future work and conclusion are presented in Sections~\ref{sec_future_work} and  \ref{sec_conclusion} respectively.

\section{Background on Mathematics for Kernel Measures}
\label{sec_background_math}
This section reviews the mathematical foundations of kernels, providing the basis for the introduction of kernel measures in the next section.
%Following Mercer's theorem a positive semi-definite kernel function implies a mapping to a Hilbert space $\hilspace$ \cite{mercer1909xvi}.
For any dataset $\set{X}$, the kernel function $\mathcal{k}: \set{X} \times \set{X} \rightarrow \mathbb{R}$ implicitly defines a feature mapping $\phi:\set{X} \rightarrow \hilspace$ that lifts the data from the input space to a `new' feature space $\hilspace$ following $\mathcal{k}(\ve{x}, \ve{x}')= \langle  \phi(\ve{x}), \phi(\ve{x}')\rangle$.
%This is what makes kernels a valuable tool in machine learning applications, they allow for the design and selection of the implicit feature space to better represent the data for the given task \cite{hofmann2008kernel, du2024exploring}.
By Mercer's theorem, any positive semi-definite kernel function implies a mapping to a Hilbert space $\hilspace$ \cite{mercer1909xvi}, so the feature map $\phi$ need not be explicitly defined. 
Decisions can be based solely on the inner product of pairwise combinations of data points. 
This is often referred to as the \forceemph{kernel trick}. 
%, so the absolute position of a individual point in the feature space $\phi(\ve{x})$ is unnecessary.
%A kernel can be interpreted as the inner-product or \forceemph{similarity} \forceemph{in} that Hilbert space.
On a finite dataset $\set{X}$ with $J$-many members, $\mathcal{k}$ can be represented by a $J\times J$ kernel matrix $\ve{K}$ (also called the Gram Matrix) where $K_{ij} = \mathcal{k}(\ve{x}_i, \ve{x}_j)$.

A classification function predicts the class label of a point $\ve{x}_*$.
Following the Representer theorem \cite{scholkopf2001generalized}, the optimal classifier can be represented as a linear combination of kernel evaluations
\begin{equation}
\label{eq_classification}
\begin{aligned}
\class (\ve{x}_*) &= \sign \left(f(\ve{x}_*)\right), \\ f(\ve{x}_*) &= \sum_{j = 1}^J \alpha_j \mathcal{k}(\ve{x}_*, \ve{x}_j),
\end{aligned}
\end{equation}
allowing for direct solutions, e.g.\ with regularized regression $\ve{\alpha} = (\ve{K}^\intercal \ve{K} + \lambda\ve{I})^{-1} \ve{K}^\intercal \ve{y}$. 
The solution generated by support vector machines (SVM) \cite{suthaharan2016support} also takes the form of Equation~\ref{eq_classification}; in this case the coefficients $\alpha_j$ are non-zero only for the \emph{support vectors} $\phi(\ve{x}_j)$, which are selected to maximize class separation in the Hilbert space.
%Kernels are a useful tool for machine learning task because the kernel can be designed such that the Hilbert space $\hilspace$ is suited for the task.
%For an example, consider the task of classification.
%A `good' Hilbert space for this task is one that results in a high classification accuracy. 
%when using Equation~\ref{eq_classification}, one can imagine what such Hilbert spaces could look like.
%A simple example of a `good' Hilbert space is one where all points belonging to  class have are mapped close together and points from different classes not mapped close together.
%A Hilbert space that might be equally as useful is one where data points that have been mapped close together belong to the same class, and data points from different classes are not mapped close together.
%A measure on kernel should capture this `usefulness' as it pertains to kernel performance on the given task.

The Frobenius inner product between two kernels of the same dimensions $\ve{K}, \ve{K}'$ is defined as
\begin{equation}
\label{eq_kernel_align}
\frobprod{\ve{K}}{ \ve{K}'} := \sum_{i,j = 1}^J  \hat{K}_{ij} K'_{ij}
\end{equation}
%$$\frobprod{\ve{K}}{ \ve{Y}} := \sum_{i,j = 1}^J  K_{ij} Y_{ij}$$
where $\hat{K}_{ij}$ is the complex conjugate.
As all kernel entries are real-valued in this work, the Frobenius inner product corresponds to the sum of the entries of the Hadamard product $\ve{K} \circ \ve{K}'$.
The inner product induces a norm $||\ve{K}||_F = \sqrt{\frobprod{\ve{K}}{ \ve{K}}}$ and has an associated metric on the space of all positive semi-definite $J \times J$ matrices.
To emphasise, within the context of Frobenius inner products or any measures \forceemph{on} kernels (kernel measures), the space we are measuring in is \forceemph{not} the Hilbert space created \forceemph{by a kernel function}, but rather the space of all kernels that could be created on the dataset $\set{X}$ (realized as the space spanned by all positive semi-definite $J \times J$ matrices).
Kernels \forceemph{measure} similarity \forceemph{in} the Hilbert space $\hilspace_{\phi}$, \forceemph{kernel measures} are measured in the space spanned by all possible kernels.
\begin{comment}
The \forceemph{confidence} $C_{j, y_i}$ \cite{hofer2024locally} captures the proportion of neighbours (weighted by similarity) of $\ve{x}_j$ which agree with the ground truth label $y_j$. 
A confidence matrix is constructed
$$\ve{C} = \ve{G}^\intercal \ve{K} $$
Confidence is closely linked with outlier detection~\cite{ramirez2010similarity}; if $\ve{x}_j$ is similar to points that mostly belong to a different class, $\ve{x}_j$ could be an outlier.
In cases where the confidence of a point is low the data points can be weighted less heavily in classification, which leads to higher classification accuracy overall \cite{hofer2024locally}.
But confidence can be used for more than classification.
\end{comment}

\section{Literature Review}
\label{sec_related_work_measures}
As mentioned in the introduction, kernel measures offer a practical approach to assessing kernel quality without the need for repeated training and evaluation of models.
Most kernel measures compare two kernels; in classification, this comparison is typically between a candidate kernel and a kernel encoding the \forceemph{ground truth}.
The ground truth may be encoded by the $J \times J$ `ideal' kernel matrix
$$  K_{\text{ ideal, } ij}  := \begin{cases}
1 & \mbox{ if } y_i = y_j\\
-1 & \mbox{ else.}
\end{cases}$$
or with a one-hot encoding resulting in a $J \times C$ matrix
$$
\ve{G}_{ic}  := \begin{cases}
1 & \mbox{ if } y_i = c\\
0 & \mbox{ else.}
\end{cases} $$ 
Note, under \forceemph{centring}, a technique utilized by many kernel measures introduced below, $\Kideal \approx \ve{G}\ve{G}^\intercal$. %Section~\ref{sec_related_work_measures}, 
%A $0,1$ ideal matrix can be constructed from $\ve{G}$ with  ${\Kideal}_{-0,1} = \ve{G}\ve{G}^\intercal$, this is the same as ${\Kideal}$ with all the $-1$ entries are replaced with $0$. 
%In some of the kernel measures to follow the kernels are \forceemph{centred}, and thus the definition of the ideal kernel (be it $\{-1, 1\}$ or $\{0,1\}$) does not effect the kernel measure.
%In the binary-class case, where $\ve{y}$ is a vector of $-1,1$ class labels, ${\Kideal} = \ve{y} \ve{y}^\intercal$.

A wide range of kernel measures have been proposed in the literature, many of which extend beyond classification tasks. 
%A review of kernel measures is presented below. 
These kernel measures can be grouped into alignment-based measures and non-alignment-based measures. 
Some non-alignment-based measures reduce to an alignment-based measure under particular parameter choices. %for classification applications the kernel alignment is usually measured between candidate kernel with the \forceemph{ideal} kernel, however applications where the alignment between two kernels are also discussed.

Kernel alignment was first proposed in \cite{cristianini2001kernel} 
\begin{equation}
\label{eq_kern_align}
\KA(\ve{K}) := A(\ve{K}, {\Kideal})
= \frac{\frobprod{\ve{K}}{{\Kideal}}}
       {||\ve{K}||_F ||{\Kideal}||_F }.
\end{equation}
%This was explained in section \ref{sec_kern_align}.
%The work in \cite{lanckriet2004learning} uses this kernel alignment to construct a two-stage learning algorithm for classification.
%The first stage learns the kernel $\ve{K}$ by optimizing the coefficients of a linear combination of kernels $\ve{K}_f$ to maximize the alignment of $\ve{K}$ to the ideal.
%By restricting the kernel matrix to have a trace of 1 and restrict kernel weights to non-negative, the kernel alignment becomes a quadratically constrained quadratic programming problem
%\begin{align*}
%\ve{K} &=\max \sum_{f =1}^F \beta_f \frobprod{\ve{K}_f}{ \ve{Y}} \\
%& s.t. \ve{\beta} \in \mathbb{R}^F_+, \sum_{f = 1}^F \sum_{h =1}^F \beta_f \beta_h \frobprod{\ve{K}_f}{\ve{K}_h} \leq 1.
%\end{align*}
%The second stage learns the support vector machine (SVM) classifier.
%It is reported in \cite{lanckriet2004learning}
Kernel alignment is sensitive to the mean of the kernel \cite{cortes2009learning}. 
As a result, kernels with larger means can obtain higher alignment scores even if they are not better suited for the given task. 
%Centered kernel alignment (CKA) addresses this limitation by removing the mean component prior to computing the alignment.

\forceemph{Centred Kernel Alignment} (CKA) was introduced in \cite{cortes2009learning} to address this limitation by removing the mean component prior to computing the alignment.
Since its introduction, CKA has given rise to a substantial body of work, with studies exploring both its theoretical properties and its practical utility across a wide range of machine learning tasks \cite{wang2015overview,ramona2012multiclass, lu2014multiple, islam2025kernel, cardenas2016centered, tian2023continuous, ding2019identification, zhong2013optimizing, alvarez2022gaussian}.
%A good overview of these works can be found in \cite{wang2015overview}.
In CKA, both of the kernels to be evaluated must first be centred with
\begin{align*}
\mathcal{k}_{\text{cent}}(\ve{x},\ve{x}') = & \langle \phi(\ve{x}) - E_x[\phi], \phi(\ve{x}') - E_{x'}[\phi] \rangle\\
= & \mathcal{k}(\ve{x},\ve{x}') - E_x[\mathcal{k}(\ve{x},\ve{x}')] - E_{x'}[\mathcal{k}(\ve{x},\ve{x}')] \\ & + E_{x, x'} [\mathcal{k}(\ve{x},\ve{x}')].
\end{align*} 
In matrix form, this is
$$(K_{\text{cent}})_{ij} = K_{ij} - \frac{1}{J} \sum_{g = 1}^J K_{gj} - \frac{1}{J} \sum_{h = 1}^J K_{ih}  + \frac{1}{J^2} \sum_{g,h = 1}^J K_{gh}.  $$
%\hl{In symmetric kernels $ \frac{1}{K} \sum_{i = 1}^J K_{ij} = \frac{1}{J} \sum_{j = 1}^J K_{ij}$ so }
%$$(K_c)_{ij} = K_{ij} - \frac{2}{J} \sum_{g = 1}^J K_{j}  + \frac{1}{J^2} \sum_{i,j = 1}^J K_{ij} $$
The alignment between the \forceemph{centred} kernels is measured using the kernel alignment from Equation~\ref{eq_kern_align} 
\begin{equation}
\label{eq_cka}
\begin{aligned}
\CKA(\ve{K})  := & A(\Kcent, \ve{K}_{\text{ideal, cent}})\\
 = & \frac{\frobprod{\Kcent}{ \ve{K}_{\text{ideal, cent}}}}
       {||\Kcent||_F || \ve{K}_{\text{ideal, cent}}||_F }.
\end{aligned}
\end{equation}
The computational complexity of centering is $O(J^2)$, and computing the alignment between the centred kernels is also $O(J^2)$. 
Thus, the overall computational complexity of CKA is $O(J^2)$. 
CKA is invariant to both orthogonal transformation and isotropic scaling. %($A(\Kcent, \ve{K}_{\text{ideal, cent}}) = A(\Kcent \ve{U}, \ve{K}_{\text{ideal, cent}} \ve{V})$ for orthonormal $\ve{U}, \ve{V}$ )  ($A(\Kcent,\ve{K}_{\text{ideal, cent}}) = A(a \Kcent, b\ve{K}_{\text{ideal, cent}} )$ for $a,b \geq 0$) \cite{kornblith2019similarity}.

%CKA also has use cases beyond use as a target function when learning linear combinations of kernels.
%In recent years CKA has seen vast applications beyond use as a target function for learning combinations of kernels.
In classification problems, CKA has been used to tune kernel parameters \cite{zhong2013optimizing, tian2015efficient, alvarez2022gaussian} and for feature selection \cite{ramona2012multiclass}.
In \cite{tian2023continuous}, the CKA is used to measure the loss for incremental learning of a neural network for classification.
However, CKA has applications beyond classifications as well.
In neural network research, CKA has been used to study representational similarity and network interpretability \cite{kornblith2019similarity, vance2024refining}, to analyse learned kernel representations \cite{canatar2022kernel, kang2026spectral, liu2026humanguidenet}, and for architectural refinement \cite{hu2024mpruner}.
CKA has been applied to molecular feature extraction \cite{wong2009constructive}, multiple kernel clustering \cite{lu2014multiple}, and for multiple information integration in medical research \cite{ding2019identification}.
 %CKA is used for network interpretability in \cite{kornblith2019similarity, vance2024refining}, and for architectural refinement in \cite{hu2024mpruner}.
%In \cite{wong2009constructive}, CKA is used for feature extraction of molecular structure. %to measure the similarity between layers of a neural network concluding that neural networks constructed with different architectures and different initialisation have similar layers. 
%In it is used to evaluate kernels learned by neural networks. 
%CKA is also used for kernel clustering in \cite{lu2014multiple}, or for multiple information integration in medical research in \cite{ding2019identification}.

CKA is not without known shortcomings: by favouring kernels that assign high similarity to all pairs of samples from the same class, it can neglect important local structure in the data \cite{wang2017multiple}.
The work in \cite{wang2017multiple} presents a solution to this in the form of a `Hybrid' Kernel Alignment (HKA), a kernel measure which combines kernel alignment at the local and global scale.
This work also presents an algorithm for learning a kernel which optimizes the HKA. 

To measure HKA, the two kernels to be evaluated must be centred.
The neighbourhood of $\ve{x}_i$ is defined as the set $\mathcal{N}_k(\ve{x}_i)$ of the $k$ nearest points to $\ve{x}_i$, where both the number of neighbours k and the choice of distance metric are parameters that must be selected or tuned. 
A local kernel is defined for the neighbourhood of each $\ve{x}_i$
$$K^{(i)}_{jk} =
\begin{cases}
(K_{\text{cent}})_{jk} & \text{if } \ve{x}_j, \ve{x}_k \in \mathcal{N}_k(\ve{x}_i)\\
0 & \text{else}.
\end{cases} $$
The local kernel alignment (LKA) between a kernel and a local centred ideal kernel is
\begin{equation}
\label{eq_LKA}
\begin{aligned}
\LKA(\ve{K}) := & A_l(\Kcent, \ve{K}_{\text{ideal, cent}}) \\ = & \frac{1}{J}\sum_{i = 1}^J \left( \frac{\frobprod{\ve{K}^{(i)}}{ {\Kideal}^{(i)}}}{ ||\ve{K}^{(i)}||_F ||{\Kideal}^{(i)}||_F} \right).
\end{aligned}
\end{equation}
Finally, the local kernel alignment is combined with the centred (global) kernel alignment following \cite{cortes2009learning} with
\begin{equation}
\label{eq_HKA}
\HKA(\ve{K}) := (1 - \lambda) \LKA(\ve{K}) + \lambda  \CKA(\ve{K}).
\end{equation}
Thus, HKA requires the selection or tuning of the size of neighbourhood $k$ and of the trade-off parameter $\lambda$.
In \cite{wang2017multiple}, this is performed empirically by fixing one parameter and varying the other to evaluate performance, with choice of parameters varying the final accuracy by up to 10\% on their benchmark datasets.
%While kernel alignment, CKA, and HKA were designed to be learn kernel combinations in MKL, they have many diverse usage, as good overview of which can be found in \ref{wang2015overview}.
%Measures of kernel alignment are used in feature extraction \cite{wong2009constructive}, feature selection \ref{ramona2012multiclass}, and kernel parameter learning \cite{keerthi2003asymptotic}.

%The Person Correlation Coefficient (PCC) is typically used to measure the correlation between $x,y$ pairs of data
%$$ \textrm{PCC} = \frac{\sum_{n = 1}^N (x_i - \bar{x})(y_i - \bar{y})}{\sqrt_{n =1}^N (x_i - \bar{x})^2} {\sqrt_{n =1}^N (y_i - \bar{y})^2} }.$$

The Hilbert-Schmidt Independence Criterion (HSIC) is a kernel measure used in some areas of research, such as neuroscience, that measures the difference between two distributions (represented by kernels) %is another kernel measure
\begin{align*}
\HSIC(\ve{K},\ve{K}' ):= & \trace(\ve{K}\ve{K}') \\ = & \sum_{i = 1}^J \sum_{j = 1}^J K_{ij} K'_{ij} = \frobprod{\ve{K}}{ \ve{K}'}.
\end{align*}
%\hl{Kernel alignment can be interpreted as a normalized version of HSIC}.
This is the non-normalized foundational metric upon which CKA is built. 
HSIC measures the statistical independence between two variables by computing the norm of the cross-covariance operator in Reproducing Kernel Hilbert Spaces (RKHS). 
While HSIC calculates the absolute dependence, CKA is HSIC normalized to stay strictly between 0 and 1. 
%Note that $\trace(\bar{\ve{K}}_1 \bar{\ve{K}}_2)= 
%The trace of the kernel products is equivalent to the Frobinius inner product 
Representational Similarity Analysis (RSA) \cite{kriegeskorte2008representational} is a measure using the HSIC and taken between two `representational dissimilarity matrices' (RDM), which are a special case of kernel. 
The work in \cite{williams2024equivalence} shows that if that kernel measure is the cosine similarity, and if the RDMs are \forceemph{centred}, then RSA is equivalent to CKA
$$\textrm{RSA}(\ve{K}, \ve{K}') := \frac{\trace(\ve{K}_{\text{cent}}\ve{K}'_{\text{cent}})}{||\ve{K}_{\text{ cent}}||_F ||\ve{K}'_{\text{ cent}}||_F }.$$ 

Canonical Correlation Analysis (CCA) and its variants, including singular vector CCA \cite{raghu2017svcca} and projection-weighted CCA \cite{morcos2018insights}, are kernel measures to measure similarity between deep neural network representations.
The work in \cite{kornblith2019similarity} shows that these methods are closely related to CKA and demonstrates through experiments that CKA is a strong measure of similarity between deep neural network layers. 
%Note that in \cite{kornblith2019similarity}, the CKA is used to compare two layers, represented as kernels, and not to compare a kernel to an ideal kernel, the ideal kernel cannot be defined in non-classification based tasks.

Feature Space-based Kernel Matrix Evaluation Measure (FSM) proposed in \cite{nguyen2007kernel} is a kernel measure designed to capture the separability of the classes of data is in the kernel space using the class means and within-class variances. 
%This measure has the same computation time as the alignment to the ideal kernel $O(n^2)$ so it is quick to compute. 
Additionally, Kernel Class Separability (KCS) \cite{xiong2005optimizing} is based on the Fisher separability criterion (FSC) in the kernel feature space and is defined as $ \textrm{FSC} := \frac{\trace(\ve{S}_b)}{\trace(\ve{S}_w)}$
where $\ve{S}_b  = \sum_{c = 1}^{C} \frac{1}{|\set{X}^c|}  (\ve{a}_c- \ve{a}) (\ve{a}_c - \ve{a})^\intercal$ is the between class separability matrix and $\ve{S}_w  = \frac{1}{J} \sum_{c = 1}^{C} \sum_{\ve{x} \in \set{X}^c}  (\phi(\ve{x}) -\ve{a}_c) (\phi(\ve{x}) - \ve{a}_c)^\intercal$ is the within class separability matrix, where $\ve{a}$ is the mean in the kernel feature space. 
A regularized variant of the Fisher separability criterion is
\begin{equation} 
\label{eq_rfc}
 \RFSC(\ve{K}) := \trace\left( (\ve{S}_w + \lambda \ve{I})^{-1} \ve{S}_b\right).
\end{equation}
Maximum Margin Criterion (MMC), a kernel measure presented in \cite{li2003efficient}, also captures binary class separability in the kernel feature space. 
FSM, FSC, and MMC rely on class-wise means, effectively favouring kernels that represent each class as a single cluster in the feature space.
However, this construction may be restrictive, as it can disadvantage kernels that capture other class structures in the data.
%For example \cite{wan2023global, wan2020within} many real-world datasets were found to follow a within-class multimodal distributions, and it has long since been established that Gaussian mixture models are an effective \hl{structure, approximation, what is the right term?} to use for audio, speech, and image processing \cite{reynolds2000speaker, blekas2005spatially}.
% \hl{technically if there is a kernel that maps to multiple separate kernels then in theory there would be a kernel that puts all those clusters together but we don't know how to find that, do we}.

%In applications such as classification, the kernel can be interpreted as a measure of \forceemph{similarity}; that is, $\mathcal{k}(\ve{x}, \ve{x}')$ is large when $\ve{x}, \ve{x}'$ are similar and small otherwise.
Most of the aforementioned kernel measures reward kernels following overly restrictive requirements than needed for classification. 
%the `similar $\leftrightarrow$ same class' assumption. %, meaning that they reward kernels that assign high similarity to all pairs of points belonging to the same class.
In particular, FSM, FSC, and MMC assume distributions in which each class forms a cluster, and kernel alignment techniques assume that the ideal kernel represents the ideal class structure, following the `similar $\leftrightarrow$ same class' assumption.
These measures may therefore fail to reward kernels that preserve within-class dissimilarities, which may be essential for capturing the underlying class structure of the data. 
Thus, there is an inconsistency between the objective of these kernel measures, which reward kernels for which all points from the same class are similar, and the requirements of an effective kernel for classification, which only require that highly similar points belong to the same class.
This inconsistency is only exacerbated when one considers that many real-world datasets contain multimodal class structures  \cite{wan2023global, wan2020within, hofer2022model}.
Additionally, it has long been established that Gaussian mixture models provide an effective modelling framework for audio, speech, and image processing and classifications \cite{reynolds2000speaker, blekas2005spatially}, further reflecting the prevalence of multimodal class structure in practice.

A kernel measure that does not follow this assumption can be constructed by measuring the smoothness of a graph Laplacian. 
We take a kernel matrix to be the weights of the edges of a graph $K_{ij} = W_{ij}$. 
The associated graph Laplacian is $\ve{L} = \ve{D} - \ve{W}$ \cite{chung1997spectral}. 
The Laplacian quadratic form provides a measure of smoothness, capturing the extent to which large edge weights connect points of different classes \cite{belkin2005manifold}
\begin{equation}
\label{eq_graph_laplacian}
\begin{aligned}
\GL(\ve{K}) :=& \ve{y}^\intercal \ve{L} \ve{y} \\ =& \frac{1}{2}\sum_{i = 1}^J \sum_{j = 1}^J  \mathcal{k}(\ve{x}_i, \ve{x}_j) \left(y_i - y_j \right)^2.
\end{aligned}
\end{equation}
This kernel measure follows the `different class $\rightarrow$ dissimilar' assumption, penalizing kernels that
assign high similarity to points from different classes.
A low GL suggests that there is low similarity between points of different classes. %kernel measure does not capture ssimilarity between points belonging to the same class, but rather it follows the assumption only captures similarity between points of different classes.
Within the context of classification, the difference between $y_i$ and $y_j$ reduces to 0 ($y_i$ and $y_j$ are the same label) or 1 ($y_i$ and $y_j$ are different labels). 
Thus, $ \GL(\ve{K}) :=  \frac{1}{2}\sum_{i = 1}^J \sum_{j = 1}^J  \mathcal{k}(\ve{x}_i, \ve{x}_j)(1- (K_{\text{ ideal, } 0,1})_ {ij}) = \frac{1}{2} \frobprod{\ve{K}}{ \ve{1}- {\Kideal}_{0,1}}$ where $\ve{1}$ is a $J\times J$ all ones matrix. 

There also exist kernel measures that evaluate properties other than the agreement between two kernels.
One such measure is the Rademacher complexity \cite{cortes2013learning}, which evaluates the expressive capacity of the function class induced by the kernel. 
A kernel with high Rademacher complexity can induce classifiers that accurately fit random label assignments.
This contrasts with most other kernel measures presented in this work, as it does not depend on the ground truth labels.
While this provides greater expressive power, it also increases the risk of overfitting and does not necessarily indicate that the kernel is well suited to the classification task.

\section{A New Measure on Kernels}
\label{sec_a_new_measure} 
The `similar $\leftrightarrow$ same class' assumption used by kernel-alignment-based kernel measures is an overly restrictive assumption for classification that penalizes some kernels which capture multimodal class structure of the data. 
%When Centred Kernel Alignment (from Section~\ref{sec_related_work_measures}) is used for classification, it follows the assumption that the ideal kernel represents the optimal class structure, and kernels with high alignment to the ideal will therefore create classifiers with high classification accuracy.
%However, the structure defined by the ideal kernel is an overly restrictive target for classification.
%The ideal kernel measures all pairs of points from the same class as equally similar, and all pairs of points from different as equally dissimilar.
%The classifier defined in Equation~\ref{eq_classification} only requires that the kernel induces sufficiently large similarity among pairs of points that contribute most strongly to the classification. %decision function. 
To achieve high classification accuracy (see Equation~\ref{eq_classification}), the kernel must induce sufficiently large similarity among pairs of points that contribute most strongly to the classification. 
Thus, the less restrictive assumption of `similar $\rightarrow$ same class' is a \forceemph{better} assumption for classification.

Consider a dataset of size $J = 8$ comprising two balanced, multimodal classes. 
The ideal kernel and a minimal-information kernel, denoted $\Kmin$, are defined below; the latter contains only the information necessary to achieve perfect leave-one-out classification accuracy 
%This kernel represents a radial kernel with a small bandwidth such as the variable-bandwidth Gaussian kernel in Equation~\ref{eq_guassian} from future section~\ref{sec_eval}. 
%The small bandwidth results in each point only being similar to it's nearest neighbour, which in this case is a symmetric relationship
\begin{align*}
\Kideal= {\footnotesize\begin{pmatrix}
 1 & 1 & 1 & 1 & 0 & 0 & 0 & 0\\
 1 & 1 & 1 & 1 & 0 & 0 & 0 & 0\\
 1 & 1 & 1 & 1 & 0 & 0 & 0 & 0\\
 1 & 1 & 1 & 1 & 0 & 0 & 0 & 0\\
 0 & 0 & 0 & 0 & 1 & 1 & 1 & 1 \\
 0 & 0 & 0 & 0 & 1 & 1 & 1 & 1 \\
 0 & 0 & 0 & 0 & 1 & 1 & 1 & 1 \\
 0 & 0 & 0 & 0 & 1 & 1 & 1 & 1 
\end{pmatrix}} ,\\ \\
 \Kmin = {\footnotesize \begin{pmatrix}
1 & 1 & 0 & 0 & 0 & 0 & 0 & 0\\
1 & 1 & 0 & 0 & 0 & 0 & 0 & 0\\
0 & 0 & 1 & 1 & 0 & 0 & 0 & 0 \\
0 & 0 & 1 & 1 & 0 & 0 & 0 & 0 \\
0 & 0 & 0 & 0 & 1 & 1 & 0 & 0  \\
0 & 0 & 0 & 0 & 1 & 1 & 0 & 0  \\
0 & 0 & 0 & 0 & 0 & 0 & 1 & 1 \\
0 & 0 & 0 & 0 & 0 & 0 & 1 & 1
\end{pmatrix}}. 
\end{align*}
%$$ve{K}_{\mbox{ideal}}= \begin{pmatrix}
%1 & 0 & 0 & 0 & 0 & 0 \\
%0 & 1 & 0 & 0 & 0 & 0 \\
%0 & 0 & 1 & 0 & 0 & 0 \\
%0 & 0 & 0 & 1 & 0 & 0 \\
%0 & 0 & 0 & 0 & 1 & 0\\
%0 & 0 & 0 & 0 & 0 & 1 
%\end{pmatrix}.$$
The minimal-information kernel has the property of `similar $\rightarrow$ same class' but not the property of `similar $\leftrightarrow$ same class'.

When classifying following Equation~\ref{eq_classification}, both $\Kideal$ and $\Kmin$ can be used to construct a classifier with $100\%$ leave-one-out accuracy. %it would be reasonable to consider them both `useful' kernels for classification. 
The ideal kernel has a CKA of 1 (trivially), but $\Kmin$ has a CKA of $\frac{2}{\sqrt{(2J-4)}}$ (for $J = 8$ this is only 0.577). %$\frac{8}{\sqrt{8} \sqrt{36}} \approx 0.471$.
Furthermore, although $\Kmin$ can be used to achieve perfect classification accuracy for any number of points, as $J$ approaches infinity, the CKA of $\Kmin$ approaches $0$.

It should be acknowledged that $\Kmin$ could achieve a HKA of 1 with the selection of $k = 2$ (the neighbourhood size) and $\lambda = 0$ (the trade-off between local and global structure). 
However, if the kernel measure itself must also be tuned, then the assessment of usefulness becomes dependent on additional dataset-specific parameters. 
%In kernel selection or parameter tuning, a kernel measure should indicate whether a kernel is likely to be useful for classification or other machine learning task.
In this case, the evaluation procedure effectively returns to the problem of kernel design (such as choosing the neighbourhood size) rather than independently assessing kernel quality.
 
%Intuitively, this example demonstrates why the ideal kernel might not capture the intrinsic class structure of the data, and why learning or selecting a kernel that aligns closely to the ideal could prevent the discovery of other `useful' kernels, kernels which might better capture the class structure of the data.
%This structure defined by the ideal kernel is not required to achieve high kernel classification accuracy when using Equation~\ref{eq_classification}, by this equation, the kernel needs only to measure \forceemph{more} pairs of points from the same class as to have a total sum of similarity higher than pairs of points from different class \hl{this depends on the value of $\ve{\alpha}$ too for example the pairs points from the same class and that have big $\alpha$ coefficients should have a high similarity, if the $\alpha$ coefficients we don't really care what classes the points are from or what their similarity is, but this is complicated to explain in equation form}.
 
\subsection{The Confidence Score}
\label{sec_confidence}

We present the \forceemph{Confidence Score}, which conceptually measures the alignment between the ground truth and the projection of the ground truth induced by the hollowed normalized kernel. The normalization and hollowing processes are as follows. %as a measure on a kernel $\ve{K}$ to capture overall for each point $\ve{x}_i$ how many of the points \forceemph{similar} to $\ve{x}_i$ share the class of $\ve{x}_i$.

\paragraph{Symmetric Normalization} %Dense regions of highly similar points can dominate the Confidence Score, causing kernels capturing dense regions to receive artificially high scores. 
%To mitigate this effect, 
The kernel is first symmetrically normalized with $\Knorm = \ve{D}^{-\frac{1}{2}} \ve{K} \ve{D}^{-\frac{1}{2}}$, where $\ve{D}$ is a diagonal matrix for which $D_{ii} = \sum_{j = 1}^J K_{ij}$. 
Symmetric normalization scales the similarity between two points according to the relative local density of both points while preserving symmetry.
Symmetric normalization preserves the intrinsic manifold geometry \cite{coifman2006diffusion}, and prevents points in dense regions from disproportionately influencing the alignment measure due solely to differences in local sampling density.

%It rescales the geometry so every single point sits on a unit sphere in the RKHS.} 
The largest eigenvalue of a symmetric normalized kernel is 1. 
Consequently, the induced linear operator is non-expansive in the Frobenius norm, $||\Knorm\ve{A}||_F \leq ||\ve{A}||_F$.
%prevent mappings with a large norm from receiving a higher Confidence Score, even if they do not better capture the class structure.
%Symmetric normalization scales the similarity between two points based on the relative local density of both participants while preserving symmetry. %and to reduce the influence of dense regions of highly similar points in the Hilbert Space}.
%This follows the decomposition used by graph Laplacians (Equation~\ref{eq_graph_laplacian}) where the Laplacian matrix is decomposed into a diagonal matrix $D$ and a weight matrix $W$.

\paragraph{Hollowing} When learning parametrized kernels, there is a risk of \forceemph{kernel degeneration}: when a kernel approaches the identity matrix, the data becomes uncorrelated.
To detect this behaviour, the candidate kernel is \emph{hollowed} with $\Kmod = \Knorm - \diag(\Knorm)$. 
Hollowing removes self-similarity measures from the kernel, which do not provide information about the relationship between different samples and therefore do not contribute to classification. 
A degenerated kernel, once hollowed, becomes the zero matrix, and thus receives a Confidence Score of 0.
%\hl{If all diagonal entries of the kernel are equal to $\lambda$, hollowing is equivalent to subtracting $\lambda \ve{I}$, and can therefore be interpreted as a form of regularization. }
%As the kernels are defined to be positive semi-definite, a degenerated kernel is the only kernel to achieve a Confidence Score of 0. }
%Thus, $\Kmod$ can be interpreted as a \hl{hollow} graph adjacency matrix.
 %or as the weight matrix in a graph Laplacian. 
 
After hollowing, $\Kmod \ve{G}$ is referred to as the \emph{confidence matrix}, where $(\Kmod \ve{G})_{ic}$ measures the total similarity of sample $\ve{x}_i$ to \emph{other samples} belonging to class $c$. 
A high confidence value for sample $\ve{x}_i$ and class $c$ indicates that $\ve{x}_i$ has a high cumulative similarity to samples belonging to class $c$.  
The Confidence Score is then obtained by measuring the alignment between the confidence matrix and the ground-truth matrix using the Frobenius inner product, providing an overall measure of how closely the confidence matrix agrees with the ground truth matrix. 
Unlike the CKA, which measures the alignment between the two $J \times J$ kernels $\frobprod{\Kcent}{ (\ve{G}\ve{G}^\intercal)_{\text{cent}}}$, the Confidence Score measures the alignment in a $J \times C$ space.

\begin{definition}[Confidence Score]
Given a candidate kernel $\ve{K}$ and the 1-hot encoding $\ve{G}$ of the ground truth class labels, the Confidence Score (CS) is defined as
%& \frac{\frobprod{\Kmod\ve{G}}{\ve{G}}}{ ||\ve{G}||_F ||\Kmod \ve{G}||_F} \\
%\begin{aligned}
\begin{equation} 
\label{eq_conf_score}
\CS(\ve{K}) := \frac{ \frobprod{\Kmod \ve{G} }{\ve{G}}}{\sqrt{J}|| \Kmod \ve{G}||_F } 
\end{equation}
and $0$ if $\frobprod{\Kmod \ve{G} }{\ve{G}} = || \Kmod \ve{G}||_F = 0$. 
%\end{aligned}
% \frac{1}{J}\sum_{j= 1}^J \sum_{c = 1}^C G_{j,c} C_{j,c} = 
%=  \frac{1}{J} \frobprod{\ve{G}^\intercal\ve{K}}{\ve{G}}.$$
\end{definition}
The algorithm efficiently computing the Confidence Score can be found in the Appendix. 
%Like CKA, the confidence score uses the Frobinius inner product to capture the level of agreement between two matrices.
%However unlike CKA which measures the agreement between the kernel matrix and the ideal kernel, consequentially requiring every that every pairwise combination of points from the same class should be highly similar to achieve a high alignment, the CS compares the $J \times C$ matrices $\ve{G}$ and $\ve{G}^\intercal \ve{K}$ and as a result only requires points that are similar to belong to the same class to achieve a high score.

 %$\frobprod{\ve{G}}{\Kmod\ve{G}}$. %between the ground truth matrix, and the projection of the ground truth matrix by the by the kernel
%The computational complexity of the CS is comparable to that of CKA and lower than that of HKA. 
Normalization and hollowing are of complexity $O(J^2)$ and computing Equation~\ref{eq_conf_score} for a fixed number of classes is also $O(J^2)$, thus the overall computational complexity is $O(J^2)$.
Therefore, the CS has the same asymptotic order as the CKA \cite{cortes2009learning}, while HKA has the additional local alignment cost associated with the computation of the local neighbourhood kernel alignment for each point. %is the most appropriate comparable}, as both follow the `similar $\rightarrow$ same class' assumption. 
Note that, of HKA and CS, only CS does not require parameter selection.
It is well known that kernel methods, due to their $O(J^2)$ complexity, do not scale well.
To use kernel methods on large datasets, the data is usually partitioned into smaller subsets to make the computation tractable.
This method will be applied to large datasets in Section~\ref{sec_tuning_bandwidth}.
 %both measures are designed to allow kernels that preserve multimodal class structure to achieve high scores. 
%Of the two, only CS does not require parameter selection.
%Compared to CKA, the confidence score is inexpensive to compute and is a good surrogate objective function for selecting kernels or learning kernel parameters. 
%Furthermore, it can easily verify as a kernel approaches the identity matrix (degeneration) $\CS(\ve{K})$ approaches $0$. 
%However the method of kernel normalization equation~\ref{eq_normalize_kern}) results in each row of the normalized kernel having a sum of approximately, but not exactly, 1, and thus $|| \ve{K} \ve{G}||_F \neq \sqrt{J}$.
\subsection{Error Bounds}
To provide theoretical guarantees for the proposed Confidence Score, we derive an algebraic bound on the reconstruction error.
In this section, we also describe the relationship between CS and normalized reconstruction error. 
\paragraph{Algebraic Bounds}
\label{sec_alg_bounds}

For a PSD kernel matrix with maximum eigenvalue one, the discrepancy between the ground truth and its projection by the normalized kernel is bounded in terms of the Confidence Score.

%Let $G$ the  ${J \times C}$ matrix represent a ground truth target matrix containing $J$ samples across $c$ classes, where each column denotes a distinct class vector. Let $K \in \mathbb{R}^{n \times n}$ be a symmetric, positive semi-definite (PSD) kernel matrix acting on a Reproducing Kernel Hilbert Space (RKHS).
The symmetric normalization $\Knorm = \ve{D}^{\frac{1}{2}} \ve{K} \ve{D}^{\frac{1}{2}}$ ensures that $\lambda_{\max}(\Knorm) = 1$. 
%ensures that its eigenvalues lie in the interval $[0,1]$, with the largest eigenvalue equal to one. %, i.e.,$.
%Since $\ve{K}$, and therefore $\Knorm$, are positive semi-definite, the entire eigenvalue spectrum $\lambda_i$ satisfies . 
%\hl{Hollowing the normalized kernel does not preserve positive semi-definiteness; however, the eigenvalues of $\Kmod$ remain bounded in magnitude by one, such that $\max_i | \lambda_i(\Kmod) | \leq 1$.
%Thus, although $\Kmod$ is not necessarily positive semi-definite, its spectral norm remains bounded by one.
Under these conditions, we derive an upper bound on the squared reconstruction error between ground truth $\ve{G}$ and its projection by the normalized kernel $\Knorm \ve{G}$. % $||\ve{G} - \Knorm \ve{G} ||_F^2$. }
The reconstruction error metric $\mathcal{E}$ is defined via the Frobenius norm. 
Expanding the Frobenius norm gives
\begin{align}
\label{eq_exact_identity}
    \mathcal{E} &= || \ve{G} - \Knorm \ve{G} ||_F^2 \nonumber \\
    &= ||\ve{G}||_F^2 - 2 \frobprod{\ve{G}}{ \Knorm \ve{G} } + ||\Knorm \ve{G} ||_F^2 \nonumber\\
    & = J - 2  \frobprod{\ve{G}}{ \Knorm \ve{G} } + ||\Knorm \ve{G}||_F^2.
\end{align}

%\hl{By isolating the inner product from our target normalized alignment definition in Equation} \ref{eq_conf_score}, we state:
%\begin{equation}
 %  \frobprod{\ve{G}}{ \Kmod \ve{G} } = \CS(\ve{K}) \cdot \sqrt{J} ||\Kmod \ve{G}||_F.
%\end{equation}
%\hl{Substituting this directly back into our expanded error equation} (Equation \ref{eq:expanded_error}) \hl{yields the exact geometric identity:}
%\hl{And thus,}
%\begin{equation}\label{eq:exact_identity}
 %   \mathcal{E} = J - 2 \CS(\ve{K}) \sqrt{J} ||\Kmod \ve{G}||_F + ||\Kmod \ve{G}||_F^2.
%\end{equation}

Since $\Knorm$ is symmetric positive semi-definite with $\lambda_{\max}(\Knorm)=1$, its spectral norm satisfies $\|\Knorm\|_2=1$. 
Consequently, multiplication by $\Knorm$ is a non-expansive linear operator on the Frobenius norm, thus $||\Knorm \ve{G}||_F \leq ||\ve{G}||_F = \sqrt{J}$ and $||\Knorm \ve{G}||_F^2 \leq  J$. 
This property is used to bound the final term in Equation~\ref{eq_exact_identity}
%\hl{Since $||\Knorm \ve{G}||_F  \leq \sqrt{J}$, 
%\hl{Substituting this bound into} Equation~\ref{eq)exact_identity}
\begin{align*}
\mathcal{E} & \leq J - 2 \frobprod{\ve{G}}{ \Knorm \ve{G} } + J\\
	& = 2J-2\frobprod{\Knorm\ve{G}}{\ve{G}}.
\end{align*}

Matrix $\ve{W}=\diag(\Knorm)$ is the diagonal matrix containing the self-similarities, so that $\Knorm = \Kmod + \ve{W}$. 
Consequently, $\frobprod{\Knorm\ve{G}}{\ve{G}} = \frobprod{\Kmod\ve{G}}{\ve{G}} + \frobprod{\ve{W}\ve{G}}{\ve{G}}$, and  
$$\mathcal{E} \leq 2J-2\frobprod{\Kmod\ve{G}}{\ve{G}}  - 2\frobprod{\ve{W}\ve{G}}{\ve{G}}.$$
Substituting the definition of the Confidence Score, $\frobprod{\ve{G}}{ \Kmod \ve{G} } = \CS(\ve{K}) \sqrt{J} ||\Kmod \ve{G}||_F$, and the identity $\frobprod{\ve{W}\ve{G}}{\ve{G}} = \trace(\Knorm)$, bounds the final error in terms of the CS
$$\mathcal{E} \leq 2J-2\CS(\ve{K}) \sqrt{J} ||\Kmod \ve{G}||_F  - 2\trace(\Knorm).$$
Thus, the reconstruction error is upper bounded by a quantity that depends explicitly on the Confidence Score, together with the magnitude of the hollowed Confidence matrix and the self-similarity contribution.

\begin{comment}
Let $A= \sqrt{J}\|\Kmod \ve{G}\|_F$, thus 
$$\mathcal{E} \leq 2J-2\CS(\ve{K}) A  - 2\trace(\Knorm).$$
For $A > 0$, the inequality can be rearranged to define an upper bound on the Confidence Score:
%$$\mathcal{E} \leq 2J - 2\CS(\ve{K})\sqrt{J}\|\Kmod\ve{G}\|_F - 2\trace(\Knorm),$$
%and therefore
\begin{equation}
    \CS(\ve{K})
    \leq
    \frac{
        2J - \mathcal{E} - 2\trace(\Knorm)
    }{
        2A
    }.
\end{equation}
\end{comment}
\begin{comment}
As similarity functions are non-negative, the elements of $\Knorm$, and therefore $\Kmod$, are non-negative, as are the elements of $\ve{G}$. Thus, $0 \leq \Kmod \ve{G} \leq \Knorm \ve{G},$ which implies
$$
\|\Kmod \ve{G}\|_F
\leq
\|\Knorm \ve{G}\|_F
\leq
\sqrt{J}.
$$
Therefore, $||\Kmod \ve{G}||_F \leq \sqrt{J}$, and
$$
\mathcal{E} \leq 2J(1-\CS(\ve{K})) - 2\frobprod{\ve{W}\ve{G}}{\ve{G}}.
$$
As $\frobprod{\ve{W}\ve{G}}{\ve{G}} = \trace(\Knorm)$, the equation can be rearranged to define a lower bound for the Confidence Score
$$
\CS(\ve{K}) \geq 1 - \frac{\mathcal{E} + 2\trace(\Knorm)}{2J}.
$$
\end{comment}
%Substituting in the definition $ \CS(\ve{K})$ gives the normalized error bound
%\begin{equation}
%\|\ve{G}-\Kmod\ve{G}\|_F^2\leq
%2J-2\frobprod{\Kmod\ve{G}}{\ve{G}}.
%end{equation}
\paragraph{Relationship between CS and Normalized Reconstruction Error}

In contrast to the reconstruction error explored in \ref{sec_alg_bounds}, the \emph{normalized} reconstruction error between the \emph{modified} Confidence matrix and the ground truth can also be considered, as
$$ E = \frac{||  \Kmod \ve{G}- \ve{G}  ||_F }{||\ve{G} ||_F} = \frac{|| \Kmod \ve{G}- \ve{G}  ||_F }{\sqrt{J}}.$$
 This formulation permits a direct geometric relationship between the reconstruction error and the Confidence Score to be established.
Let $r$ denote the relative magnitude of the modified confidence matrix,
$$r = \frac{|| \Kmod \ve{G} ||_F }{||\ve{G} ||_F} = \frac{|| \Kmod \ve{G} ||_F }{\sqrt{J}}$$

Using the definition of the Confidence Score from Equation~\ref{eq_conf_score}, the squared normalized reconstruction error can be expanded as
\begin{align*} 
E^2 &= \frac{|| \Kmod \ve{G}- \ve{G}  ||_F^2 }{J}\\
  &= \frac{|| \Kmod \ve{G}||_F^2 + || \ve{G}||_F^2  - 2\frobprod{\ve{G}}{ \Kmod \ve{G}}}{J}\\
  & = r^2 + 1 -2r\CS(\ve{K})
\end{align*}
Thus, the normalized reconstruction error is related exactly to the Confidence Score and the relative magnitude $r$ of the modified Confidence matrix.
In particular, the Confidence Score captures the alignment between $\Kmod\ve{G}$ and $\ve{G}$, while $r$ accounts for their relative magnitude.

\subsection{Boundary-Case Kernels}
When selecting from a set of candidate kernels, a kernel measure should predict which kernel will produce the most accurate classifier.
In this evaluation, a set of kernels is constructed from the kernels ${\Kideal}$ and $\ve{K}_{\text{ min}}$ with the addition of the `degenerated' kernel $\ve{K}_{\text{ ident}}$ (the identity matrix).
%This experiment is performed on a set 
%This example illustrates how the Confidence Score identifies kernels that produce accurate classifiers that CKA fails to identify.
%This demonstrates that a useful kernel for classification should ensure that similar points belong to the same class (as measured by the CS), rather than requiring all points from the same class to be similar (as measured by the CKA).
%This example will illustrates how the confidence score reflects the assumption that a `useful' kernel for classification is one in which similar points belong to the same class, and how selecting kernels using the CS can improve classification accuracy compared to CKA.  

The CS, CKA, RFSC, and the leave-one-out accuracy for each of the three kernels are presented in Table~\ref{tab_example_kernel_score_symbolic}.
HKA is not included as it would require the additional tuning of $k$ and $\lambda$.
RFSC is included to provide a comparison to kernel measures designed to reward kernels for which each class of data is distributed around a class mean. 
Included are kernel measures for a dataset of size $J = 2N, N \in \mathbb{Z}^+, N \geq 2$ as well as the limits for these measures as $J\rightarrow \infty$.

Trivially, when the candidate kernel \forceemph{is} the ideal kernel, the CKA and the accuracy are equal to 1.
The Confidence Score for the ideal kernel is 
\begin{equation}
\label{eq_cs_kideal}
\CS(\Kideal) = \frac{1/3 J - 1/12C}{\sqrt{J} \sqrt{1/9 J - 5/96 C}}
\end{equation}
which converges to 1 as $J \rightarrow \infty$. 
%\hl{The ideal kernel collapses all points of the same class to a single point in the kernel feature space, the within class separability is 0, resulting in a RFSC of $\infty$.}
The kernel $\Kmin$ also achieves perfect classification accuracy for all values of J, yet this is not captured by the CKA or the RFSC as $\lim_{J \rightarrow \infty} \CKA(\Kmin) = 0$ and $\lim_{J \rightarrow \infty} \RFSC(\Kmin) = 0$. % $\CKA \neq 1$ despite also achieving an accuracy of 1.
%Furthermore, when the candidate kernel is $\Kident$, $\CKA$ is \forceemph{not} 0 despite achieving an accuracy of 0.

We show that, by construction, for all values of $J$, $\CS(\Kmin) = 1$.
After normalization and hollowing, $\Kminmod$ takes the form of a block-diagonal matrix whose diagonal blocks are $\frac{1}{4} \scalebox{0.6}{$\begin{pmatrix} 0 & 1 \\ 1 & 0 \end{pmatrix}$}$ on the diagonal.
Consequently, $\Kminmod\ve{G} = \frac{1}{4}\ve{G}$. 
Thus, $\frobprod{\Kminmod\ve{G}}{\ve{G}} = \frac{J}{4}$, and $||\Kminmod\ve{G}||_F = \frac{\sqrt{J}}{4}$. 
Substituting these into the definition of the Confidence Score gives
$$\CS(\Kmin) = \frac{\frac{J}{4}}{\sqrt{J}\frac{\sqrt{J}}{4}} = 1.$$
Additionally, in cases where $J \gg C$ , $\CS(\Kideal) \approx 1$.
Thus, among these example kernels, the Confidence Score remains correlated with classification accuracy, unlike CKA.  % and thus within these example kernels, the CS is more correlated with accuracy.}
%\hl{In cases where $J >> C$, the  $\CS(\Kmin) \approx 1$}, considering this along with $\CS(\Kident) = 0 $ all values of $J$, the CS attains the highest correlation with accuracy. 

\begin{table}[htbp]
\centering
\small
\setlength{\tabcolsep}{4pt}
\caption{The CKA from Equation~\ref{eq_kernel_align}, RFSC from Equation~\ref{eq_rfc}, CS from Equation~\ref{eq_conf_score}, and leave-one-out classification accuracy using three example kernels. The limit as $J \rightarrow \infty$ is presented in parentheses.  }
\label{tab_example_kernel_score_symbolic}
\begin{tabular}{lccccc}
\toprule
& \makecell{\textbf{CKA} \\($J \rightarrow \infty$)} &
\makecell{\textbf{RFSC} \\($J \rightarrow \infty$)} &
\makecell{\textbf{CS} \\($J \rightarrow \infty$)} &
\textbf{Acc.} \\
\midrule
$\Kideal$ & 1 \; (1) & $\frac{C-1}{\lambda}$ \; ($\frac{C-1}{\lambda}$) &  Eq~\ref{eq_cs_kideal} \; (1) & 1\\
$\Kmin$ & $\frac{2}{\sqrt{(2J-4)}}$ \; (0)& $\frac{C-1}{J}$ \; (0) & 1 \; (1)& 1\\
$\Kident$ & $\frac{1}{\sqrt{J-1}}$ \; (0) & $\frac{C-1}{J}$ \; (0) & 0 \; (0) & 0\\
\bottomrule
\end{tabular}
\end{table}

\section{Evaluation of the Confidence Score}
\label{sec_eval}
This section compares the performance of the Confidence Score to other kernel measures. 
First, a comparison of the usage of CKA and the CS to \forceemph{tune} kernel parameters is presented, followed by a comparison of their use for \forceemph{selecting} a kernel from a pool. 
Finally, two synthetic datasets are used to illustrate how kernels capture similarity and class structure, and how that relates to the kernel measures and classification accuracy.

This evaluation primarily focuses on the comparison of the Confidence Score with kernel-alignment-based measures. The regularized Fisher Separability Criterion (RFSC) was included as a representative of class-separability-based kernel measures. 
Although other kernel measures were considered, they were not included in the final numerical evaluation due to computational complexity or weak statistical correlation with classification accuracy in preliminary experiments. 

A \forceemph{family} of kernels can be obtained from a base kernel function by varying its parameters or changing metrics.
The following evaluations are performed using four families of kernels,
\begin{enumerate}
\item The variable-bandwidth Gaussian family of kernels,
\begin{equation}
\label{eq_guassian}
\mathcal{k}_b(\ve{x}_i, \ve{x}_j) = \exp \left( -\frac{d_{ij}^2}{2\sigma^2 b^2} \right)
\end{equation}
where $b$ is the tunable bandwidth and $d_{ij}$ is the distance between $\ve{x}_i$ and $\ve{x}_j$, usually Euclidean. %$ d_{ij} = ||\ve{x}_i - \ve{x}_j||$. 
This kernel is stationary as it depends only on the distance between points.
\item The Mat\'ern family of kernels, a popular choice of kernel \cite{palar2023kernel, muyskens2024identifiability, lu2025choice} 
$$\mathcal{k}_\rho(\ve{x}_i, \ve{x}_j) = \sigma^2 \frac{2^{1-\nu}}{\Gamma(\nu)} \left( \sqrt{2\nu} \frac{d_{ij}}{\rho}\right)^{\nu} K_\nu \left(\sqrt{2\nu} \frac{d_{ij}}{\rho} \right) $$
where $\Gamma$ is the gamma function, $K_\nu$ is the modified Bessel function of the second kind, and $\rho$ and $\nu$ are tunable parameters. 
To simplify the kernel computation, only three special cases of Mat\'ern kernel are considered where $\nu \in \{\frac{1}{2}, \frac{3}{2}, \frac{5}{2}\}$. 
In these cases, the Bessel function has a simple closed-form expression, and $\rho$ is the tunable parameter.
\begin{comment}
\begin{enumerate}
\item $$\mathcal{k}_{1/2}(\ve{x}_i, \ve{x}_j) = \sigma^2 e^{\frac{-d_{ij}}{\rho}},$$
\item $$\mathcal{k}_{3/2}(\ve{x}_i, \ve{x}_j) = \sigma^2 \left(1 + \frac{\sqrt{3}d_{ij}}{\rho} \right) e^{\frac{-\sqrt{3}d_{ij}}{\rho}},$$
\item $$\mathcal{k}_{5/2}(\ve{x}_i, \ve{x}_j) = \sigma^2 \left(1 + \frac{\sqrt{5}d_{ij}}{\rho} + \frac{5 d_{ij}^2}{3\rho^2}\right) e^{\frac{-\sqrt{5}d_{ij}}{\rho}},$$
\end{enumerate}
where $\sigma$ is the variance and $\rho$ is the tunable parameter called the lengthscale similar to the bandwidth.
\end{comment}
\item The Local Variable-Bandwidth Gaussian (LVBG) family of kernels
\begin{align}
s(\ve{x}_i, \ve{x}_j) = &\begin{cases}
1 & \mbox{ if } i = j\\
e^{ - \frac{ (d_{ij} - l_i) ^2}{\sigma_i^2}} &  \mbox{ if } \ve{x}_j \in  \mathcal{N}_K(\ve{x}_i)\\
0 & \mbox{ else,}
\end{cases} \label{eq_varibable_band_kernel} \\[0.3em]
\mathcal{k}(\ve{x}_i,\ve{x}_j) = & s(\ve{x}_i,\ve{x}_j) + s(\ve{x}_j,\ve{x}_i) \label{eq_symmetrization}\\
& - s(\ve{x}_i,\ve{x}_j)s(\ve{x}_j,\ve{x}_i) \notag
\end{align}
where $d_{ij}$ is a distance metric (in this work, either Euclidean or Manhattan), and $l_i$ is the distance to the first nearest neighbour of $\ve{x}_i$.
\item The Local Variable-Bandwidth Laplacian (LVBL) kernel, previously explored in \cite{hofer2024locally} and as used for dimensionality reduction in \cite{mcinnes2018umap}
\begin{align}
\label{eq_LSLDS}
 s(\ve{x}_i, \ve{x}_j) &=\begin{cases}1 & \mbox{ if } i = j\\
  e^{- \frac{ d_{ij}- l_i}{\sigma_i}} & \mbox{ if } \ve{x}_j \in  \mathcal{N}_k(\ve{x}_i)\\
  0 & \mbox{ else.}
  \end{cases}
\end{align}
where again $d_{ij}$ is a distance metric, and $l_i$ is the distance to the first nearest neighbour of $\ve{x}_i$.
The kernel is symmetrized as in equation~\ref{eq_symmetrization}.
%Again, $\sigma_i$ is the local bandwidth (computed as in the LSBGK), and $\hat{d}_{ij}$ is a distance metric (in this work either Euclidean or Manhattan).
 \end{enumerate}
Note that both the LVBG and the LVBL kernels have an automatic bandwidth selection; the local bandwidth $\sigma_i$ is computed to satisfy $\sum_{i = 1}^J s(\ve{x}_i, \ve{x}_j) = \log_2(K)$ for each neighbourhood. % and thus it is not a parameter that can be tuned.
These kernels create families by varying parameters such as the neighbourhood size and the distance metric. 
%The work in \cite{hofer2024locally} shows that these parameters have little effect on the kernel classification accuracy.

In the evaluations below, classification is performed following the regularization solver presented in Equation~\ref{eq_classification}, in Section~\ref{sec_tuning_bandwidth} SVMs are also utilized for comparison. 
All accuracy measurements are computed using leave-one-out testing, resulting in a deterministic measure of accuracy. %that does not depend on the choice of partition between the testing set and the training set. 
In addition to synthetic data, some experiments use datasets from the UCI Machine Learning Repository \cite{Dua:2017} and the MNIST dataset \cite{deng2012mnist}, the metadata for which can be found in Table~\ref{tab_dataset}.
%In leave-one-out testing one by one each point in $\set{X}$ is taken as a `test point', where the label information is removed the ground truth $\ve{G}$ and ${\Kideal}$, then the kernel measures are taken, the test point is classified and that classification is stored in the confusion matrix, and then the label information of the test point is returned to $\ve{G}$ and ${\Kideal}$.
%  from such that the kernel similarities between test point $\ve{x}_*$ and all other points are removed from the kernel, and therefore all kernel measures.
%Leave-one-out testing is used specifically to create a deterministic measure, as each point is used as the test point exactly once.

 \begin{table}[htbp]
 \centering
\caption{A summary of the datasets used in this research.}
\label{tab_dataset}
%\newcolumntype{C}{>{\centering\arraybackslash}X}
\begin{tabular}{lccc} 
\toprule
\textbf{Dataset Name} & \textbf{Size} & \textbf{Dimensions} & \textbf{Num. Classes} \\
\midrule
Banknote & 1372 & 4 & 2 \\
Bioddeg & 1055 & 41 & 2 \\
Diabetes & 768 & 8 & 2 \\
Glass & 214 & 9 & 7 \\
Heart & 270 & 13 & 2 \\
%Ionosphere & 351 & 33 & 2 \\
Iris & 150 & 4 & 3 \\
Lympho & 148 & 18 & 4 \\
Sonar & 208 & 60 & 2 \\
Vowel & 528 & 10 & 11 \\
Wine & 178 & 13 & 3 \\
Yeast & 1484 & 8 &  10 \\
Australian & 690 & 14 & 2 \\
Balance & 625 & 4 & 3\\
Haberman & 306 & 3 & 2 \\
Monks & 432 & 6 & 2 \\
Vehicle & 94 & 18 & 4 \\
MNIST & 20 000 & 784 & 10\\
\bottomrule
\end{tabular}
\end{table}

\subsection{Tuning Kernel Bandwidth}
\label{sec_tuning_bandwidth}

\begin{figure*}[t]
    \centering

    \begin{subfigure}{0.32\textwidth}
        \centering
        \includegraphics[width=\textwidth]{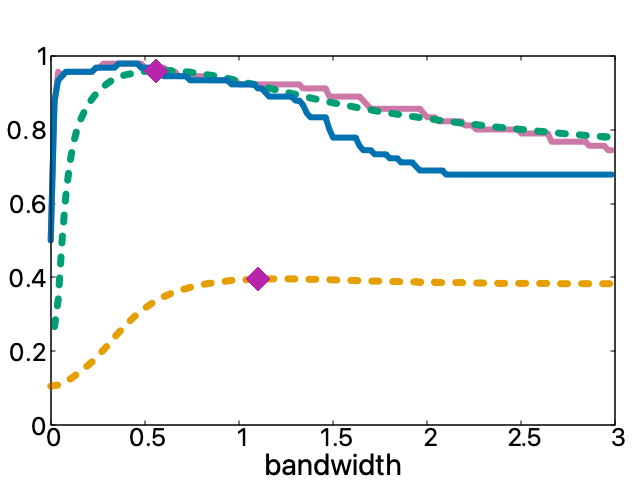}
        \caption{Clusters}
    \end{subfigure}
    \hfill
    \begin{subfigure}{0.32\textwidth}
        \centering
        \includegraphics[width=\textwidth]{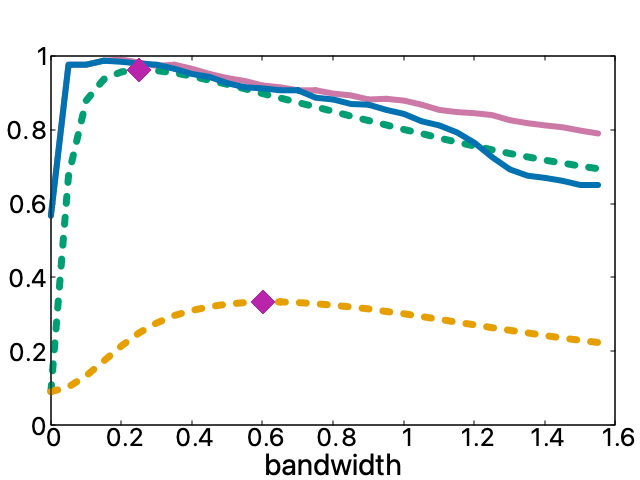}
        \caption{Moons}
    \end{subfigure}
    \hfill
    \begin{subfigure}{0.32\textwidth}
        \centering
        \includegraphics[width=\textwidth]{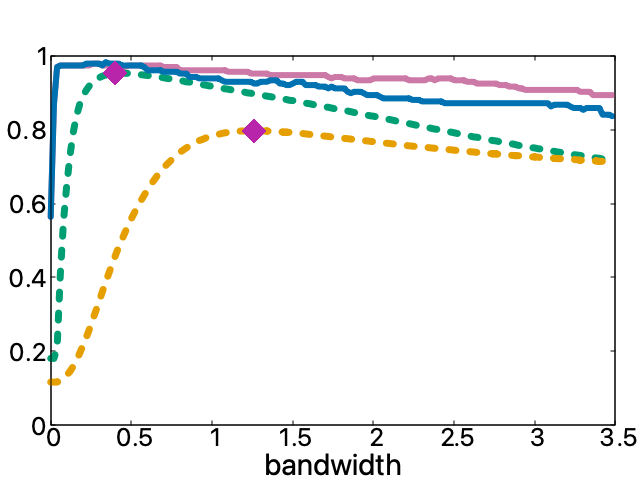}
        \caption{Iris}
    \end{subfigure}

    \vspace{2mm}

    \begin{subfigure}{0.32\textwidth}
        \centering
        \includegraphics[width=\textwidth]{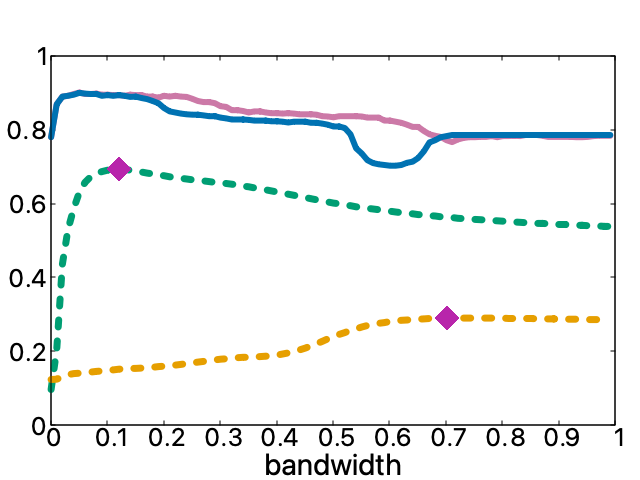}
        \caption{Glass}
    \end{subfigure}
    \hfill
    \begin{subfigure}{0.32\textwidth}
        \centering
        \includegraphics[width=\textwidth]{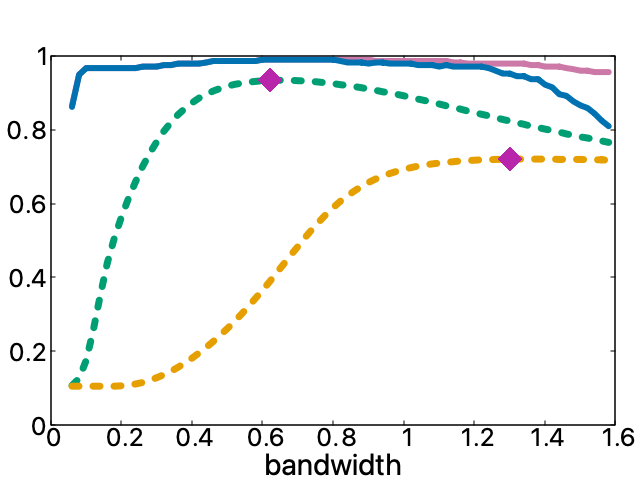}
        \caption{Wine}
    \end{subfigure}
    \hfill
    \begin{subfigure}{0.32\textwidth}
        \centering
        \includegraphics[width=\textwidth]{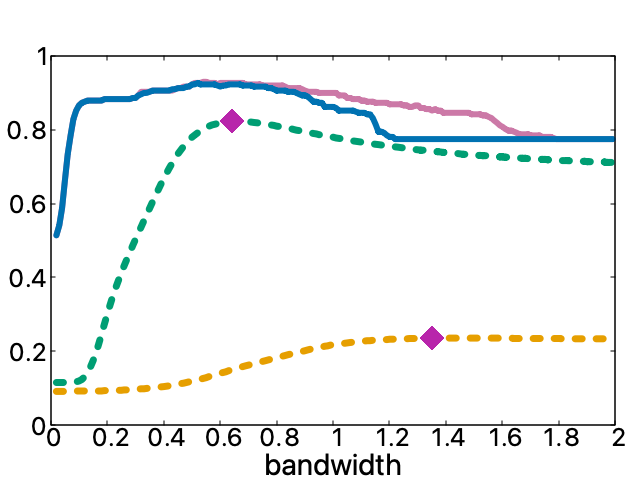}
        \caption{Lympho}
    \end{subfigure}

    \vspace{2mm}

    \begin{subfigure}{0.32\textwidth}
        \centering
        \includegraphics[width=\textwidth]{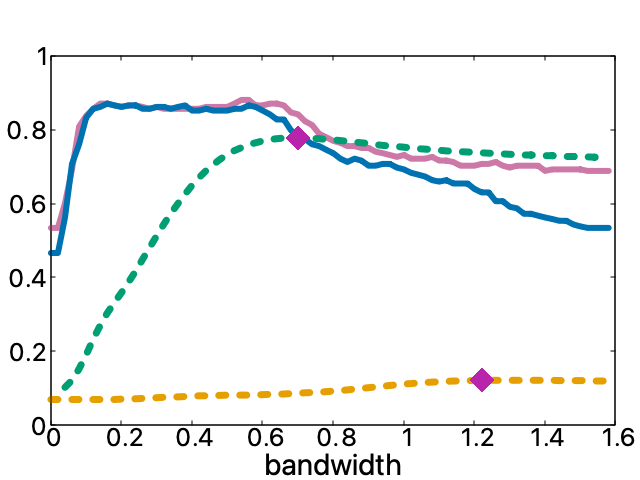}
        \caption{Sonar}
    \end{subfigure}
    \hfill
    \begin{subfigure}{0.32\textwidth}
        \centering
        \includegraphics[width=\textwidth]{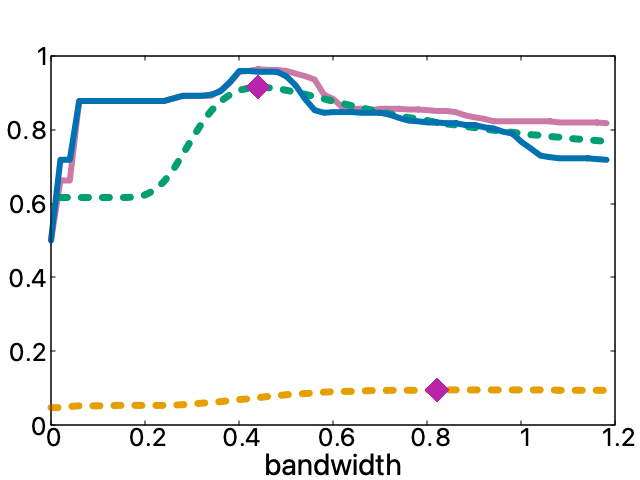}
        \caption{Monks}
    \end{subfigure}
    \hfill
    \begin{subfigure}{0.32\textwidth}
        \centering
        \includegraphics[width=\textwidth]{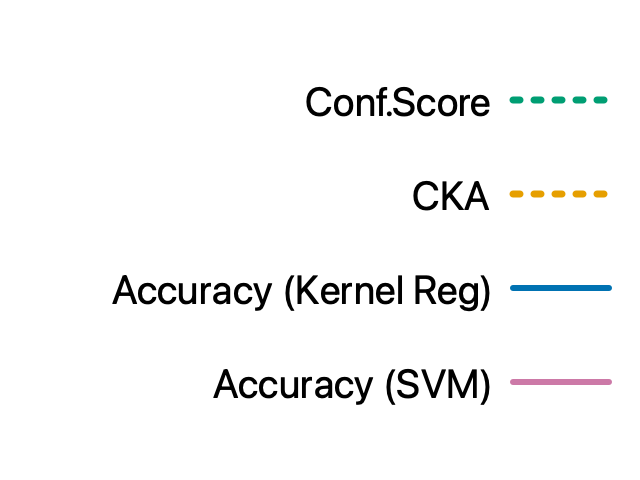}
    \end{subfigure}

    \caption{The CS, CKA, and leave-one-out classification accuracy over a range of bandwidth values on eight datasets. The maxima of the CS and CKA are denoted by purple diamonds. Since there is no unique maximum classification accuracy for most datasets, the maximum accuracy is not marked.}
    \label{fig_learn_bandwidth}
\end{figure*}

\begin{table}[htbp]
\centering
\caption{Classification accuracy (from regularized regression classifier) determined by bandwidth from maximum kernel measures compared to maximum accuracy across all bandwidths, from data presented in Fig.~\ref{fig_datasets}.}
\label{tab_accuracy}
%\rowcolors{1}{yellow!20}{yellow!20}
\begin{tabular}{lcc|c}
\toprule
\textbf{Dataset} 
& \makecell{\textbf{Accuracy at}\\\textbf{Max CKA}} 
& \makecell{\textbf{Accuracy at}\\\textbf{Max CS}} 
& \makecell{\textbf{Max}\\\textbf{Accuracy}} \\
%& \makecell{\textbf{Diff. between} \\
%\textbf{Max Acc.} \\
%\textbf{and Acc. at}
%\\\textbf{Max CS}} 
%& \makecell{\textbf{Diff. between} \\
%\textbf{Max Acc.}\\
%\textbf{and Acc. at}\\ 
%\textbf{Max CKA}} \\
\midrule
Clusters          & $0.922$ & $\best{0.967}$ & $0.978$ \\
Moons  & $0.919$ & $\best{0.981}$ & $0.986$ \\
Iris              & $0.951$ & $\best{0.973}$ & $0.982$ \\
Glass             & $0.771$ & $\best{0.893}$ & $0.899$ \\
Wine              & $0.978$ & $\best{0.989}$ & $0.989$ \\
Lympho			   & $0.774$ & $\best{0.922}$ & $0.926$ \\
Sonar             & $0.707$ & $\best{0.841}$ & $0.870$ \\
Monks             & $0.850$ & $\best{0.963}$ & $0.958$ \\
%Ionosphere        & $\best{0.892}$ & $\best{0.892}$ & $0.906$ \\
\bottomrule
\end{tabular}
\end{table}

One of the key applications kernel measures is their usage in tuning continuous kernel parameters (e.g. the bandwidth of a Gaussian kernel) \cite{keerthi2003asymptotic, camargo2009multi,zhong2013optimizing, tian2015efficient}. 
In kernel parameter tuning, the kernel with the parameter(s) that maximizes the kernel measure is expected to achieve the highest classification accuracy.
%The kernels in Figure~\ref{fig_iris_bandwidths} are constructed using a Guassian kernel with variable-bandwidth $b$, the class membership is indicated by the ideal kernel in Figure~\ref{fig_iris_ideal}.
%The kernel in Figure~\ref{fig_iris_medium_bandwidth} captures the most useful information for classification, with the kernel in Figure~\ref{fig_iris_small_bandwidth} using a bandwidth that is too small (kernel degeneration) and the kernel in Figure~\ref{fig_iris_large_bandwidth} using a bandwidth that is too large for this dataset.
%A kernel measure should identify the bandwidth for a kernel that can be used to construct the most accurate classifier.
%To assess this, the relationship between kernel measures and classification accuracy was evaluated in the context of kernel parameter tuning. 

Eight datasets were selected for this demonstration (see Table~\ref{tab_accuracy}): six from the UCI Machine Learning Repository (see Table~\ref{tab_dataset}) and two synthetic datasets (see Fig.~\ref{fig_datasets}).
For each dataset, Gaussian kernels (Equation~\ref{eq_guassian}) were constructed over a range of bandwidths, and for each kernel, the kernel measures (CKA and CS) and the corresponding classification accuracy were computed. 
The kernel measures and their maxima are presented in Fig.~\ref{fig_learn_bandwidth}, while Table~\ref{tab_accuracy} compares the maximum classification accuracy with the accuracy obtained at the bandwidth selected by each kernel measure.

\begin{figure}
    \centering
    \includegraphics[width=0.98\columnwidth]{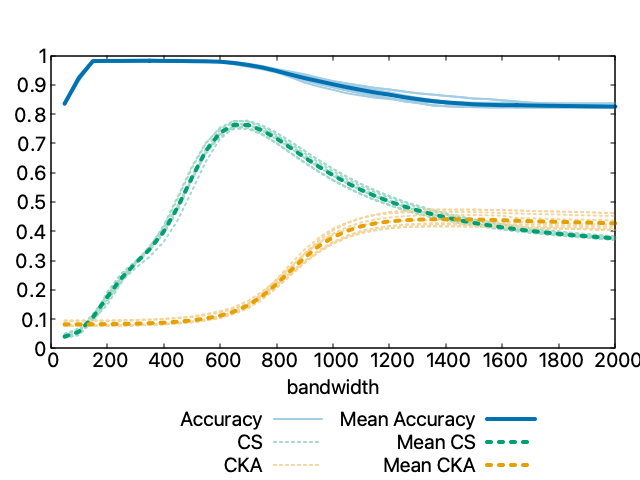}
    \caption{The CS, CKA, and leave-one-out classification accuracy of a Gaussian kernel classifier across a range of bandwidth values on 10 partitions of the MNIST dataset.}
    \label{fig_mnist}
\end{figure}

To demonstrate the capabilities of the CS on large datasets, we present the Gaussian kernel bandwidth tuning experiment on the MNIST dataset, where the data are partitioned into 10 subsets of 2000 points each. 
Figure~\ref{fig_mnist} presents the classification accuracy, CKA, and CS for each partition, as well as the mean curve for each metric. Not only does the CS consistently identify the bandwidth which produces higher classification accuracy for all partitions, but the CS varies less between partitions (average variance $1.4 \times 10^{-3}$ along curve) than the CKA (average variance $3.5 \times 10^{-3}$ along curve). 

From these experiments we observe that the CS consistently identifies a bandwidth that achieves an accuracy closer to the maximum than the CKA. %and \hl{one cases, the CS identifies a bandwidth that achieves the maximum accuracy. 
Furthermore, the trends observed in the classification accuracy in Fig.~\ref{fig_learn_bandwidth} and Fig.~\ref{fig_mnist} more closely resemble those of the CS than those of the CKA.

\subsection{Kernel Selection}
\label{sec_evaluation_kernel_selection}

Another application of kernel measures is their use for the selection of the `best' kernel from a finite pool of kernels.
This can be used to select a kernel for a kernel classifier \cite{cristianini2001kernel, chudzian2012evaluation}, as a preprocessing step in Multiple Kernel Learning (MKL) to reduce a pool of candidate kernels \cite{ramona2012multiclass, wong2011using}, or as part of the objective function in MKL \cite{kandola2002optimizing, cortes2009learning, tian2023continuous, lanckriet2004learning, kumar2012binary, lu2014multiple, wang2017multiple, wang2018two, kwok2003learning}.
For this experiment, the correlation between classification accuracy and each kernel measure is evaluated over a pool of 20 kernels consisting of:
\begin{itemize}
    \item Gaussian kernels with bandwidths \\$b \in \{0.1, 0.3, 0.5, 0.7, 0.9, 1.0, 1.5, 2.0, 2.5, 3.0\}$,
    \item Mat\'ern kernels with $\nu \in \left\{\frac{1}{2}, \frac{3}{2}, \frac{5}{2}\right\}$,
    \item three locally scaled Gaussian kernels (constructed using Euclidean distance) with number of neighbours $k \in \{5, 10, \sqrt{J}\}$, and
    \item three locally scaled Laplacian kernels (constructed using Manhattan distance) with number of neighbours $k \in \{5, 10, \sqrt{J}\}$.
\end{itemize}
This pool of kernels was constructed on 15 of the benchmark datasets from the UCI Machine Learning Repository \cite{Dua:2017} listed in Table~\ref{tab_dataset}.
%We choose this pool of kernels expecting some kernels to achieve high classification accuracy and some to achieve low classification accuracy. 

The correlation between classification accuracy and CS, CKA, and RFSC was computed across the set of datasets. 
The statistical significance of differences between the correlations was evaluated using the Cocor package for R \cite{cocor}.
The null hypothesis was that the CS is not more strongly correlated with classification accuracy than the other kernel measure (CKA or RFSC). 
Since Cocor performs several tests for comparing dependent correlations, Table~\ref{tab_correlation_p_values} reports the maximum p-value returned across all tests. 
A result was considered significant only if all tests rejected the null hypothesis at $\alpha = 0.05$.
All tested datasets in Table~\ref{tab_correlation_p_values} rejected the null hypothesis, suggesting that the CS is more strongly correlated with classification accuracy than the CKA and RFSC.

\begin{table}[htbp]
\centering
\caption{Maximum p-value obtained from the Cocor tests comparing the correlation of classification accuracy with CS against the corresponding correlations obtained with CKA and RFSC.}
\label{tab_correlation_p_values}
%\rowcolors{1}{yellow!20}{yellow!20}
\begin{tabular}{lcc}
\toprule
\textbf{Dataset} & \makecell{\textbf{p-value} \\ \textbf{CS vs CKA}} &  \makecell{\textbf{p-value} \\ \textbf{CS vs RFSC}   } \\
\midrule
Banknote     &  $0.00182$ & -- \\
Biodeg       & $0.00001$ & $0.00060$ \\
Diabetes     & $0.00000$ & $0.00917$ \\
Glass        & $0.00004$ & $0.00000$ \\
Heart        & $0.00005$ & $0.00033$ \\
%Ionosphere   & $0.555$  \\ $0.00010$
Iris         & $0.00001$ & $0.00000$ \\
Lympho       & $0.00009$ & $0.00010$ \\
Sonar        & $0.00310$ & $0.00015$ \\
Vowel        & $0.00045$ & $0.00000$ \\
Wine         & $0.00063$ & $0.00000$ \\
Yeast        & $0.00022$ & $0.00000$ \\
Australian   & $0.00970$ & $0.00037$ \\
Balance      & $0.00014$ & $0.00124$ \\
Haberman     & $0.00000$ & $0.00088$ \\
Monks        & $0.01606$ & $0.00000$ \\
%Vehicle      & $0.531$  \\
\bottomrule
\end{tabular}
\end{table}

\subsection{Capturing the Intra-Class Structure on Synthetic Multimodal Datasets}

\begin{figure}[t]
\centering
\begin{subfigure}{0.48\columnwidth}
    \centering
    \includegraphics[width=\linewidth]{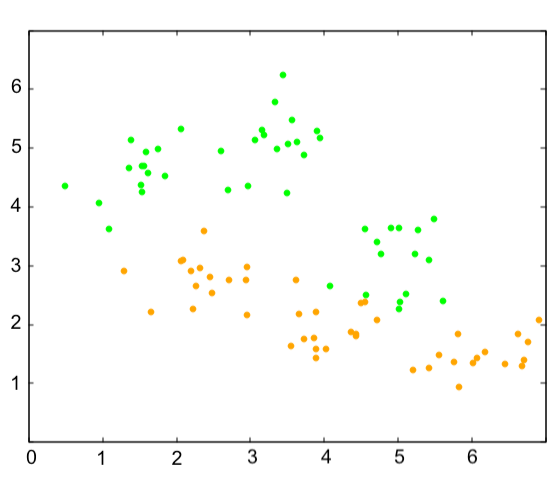}%{multi_clust_per_class}
    \caption{`Clusters' dataset.}
    \label{fig_multi_clust_per_class}
\end{subfigure}
\hfill
\begin{subfigure}{0.48\columnwidth}
    \centering
    \includegraphics[width=\linewidth]{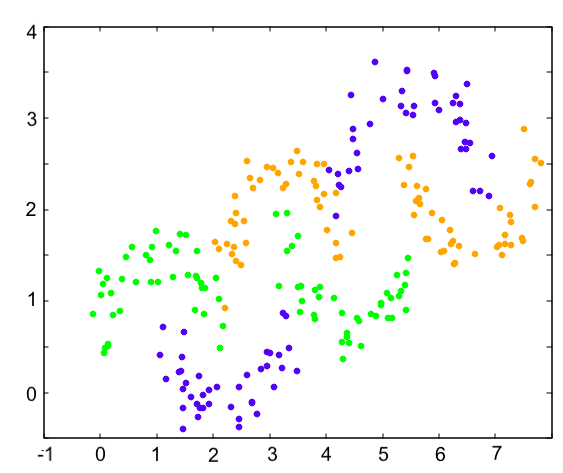}%{noisey_half_moons}
    \caption{`Noisy Crescent Moon' dataset.}
    \label{fig_noisy_half_moons}
\end{subfigure}
\caption{Synthetic datasets.}
\label{fig_datasets}
\end{figure}

\begin{figure}[htbp]
    \centering
    \begin{subfigure}{0.3\columnwidth}
        \centering
        \includegraphics[width=\linewidth]{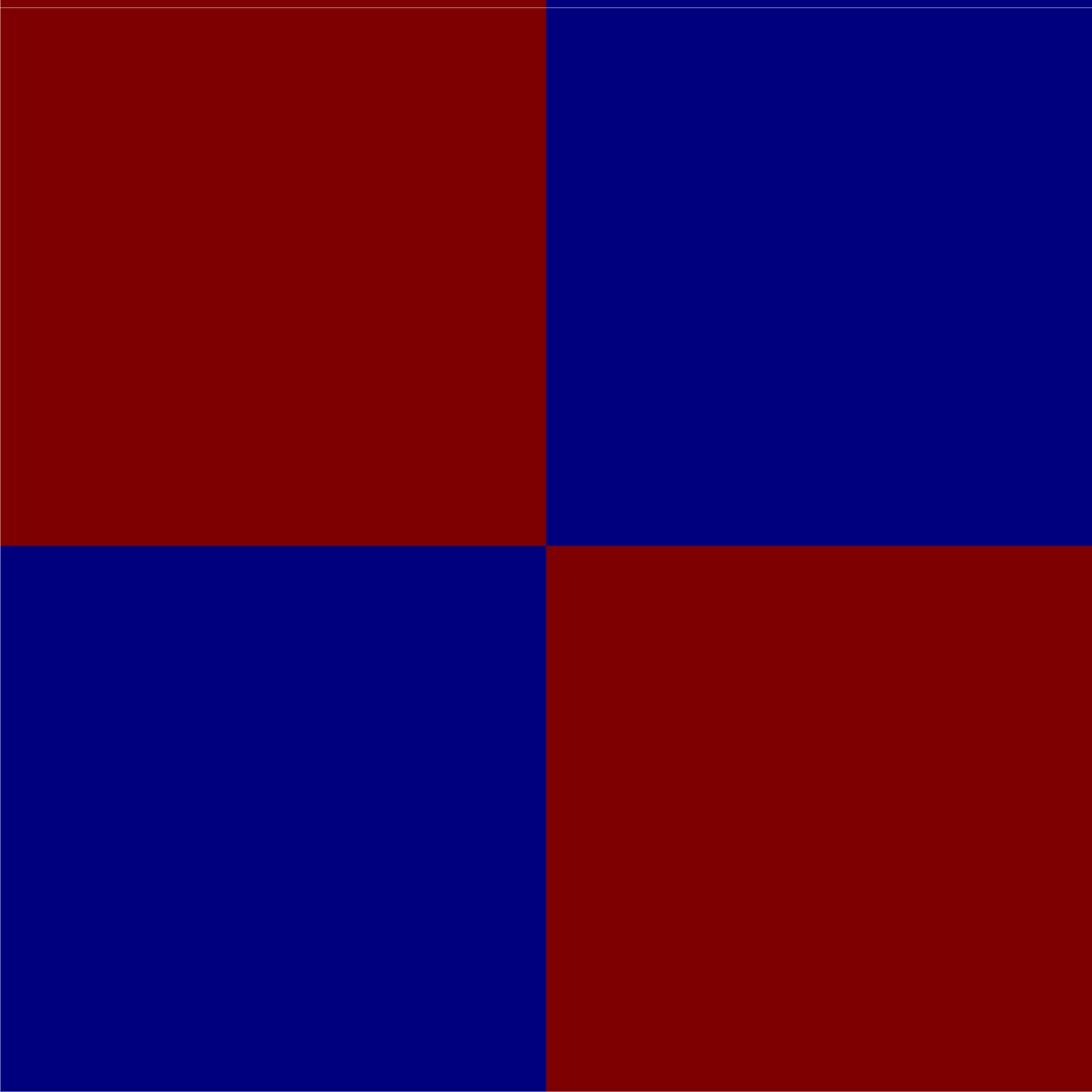}%{ideal_kernel_mutli_clust_per_class}
        \caption{ }
        \label{fig_multi_ideal}
    \end{subfigure}
    \hfill
    \begin{subfigure}{0.3\columnwidth}
        \centering
        \includegraphics[width=\linewidth]{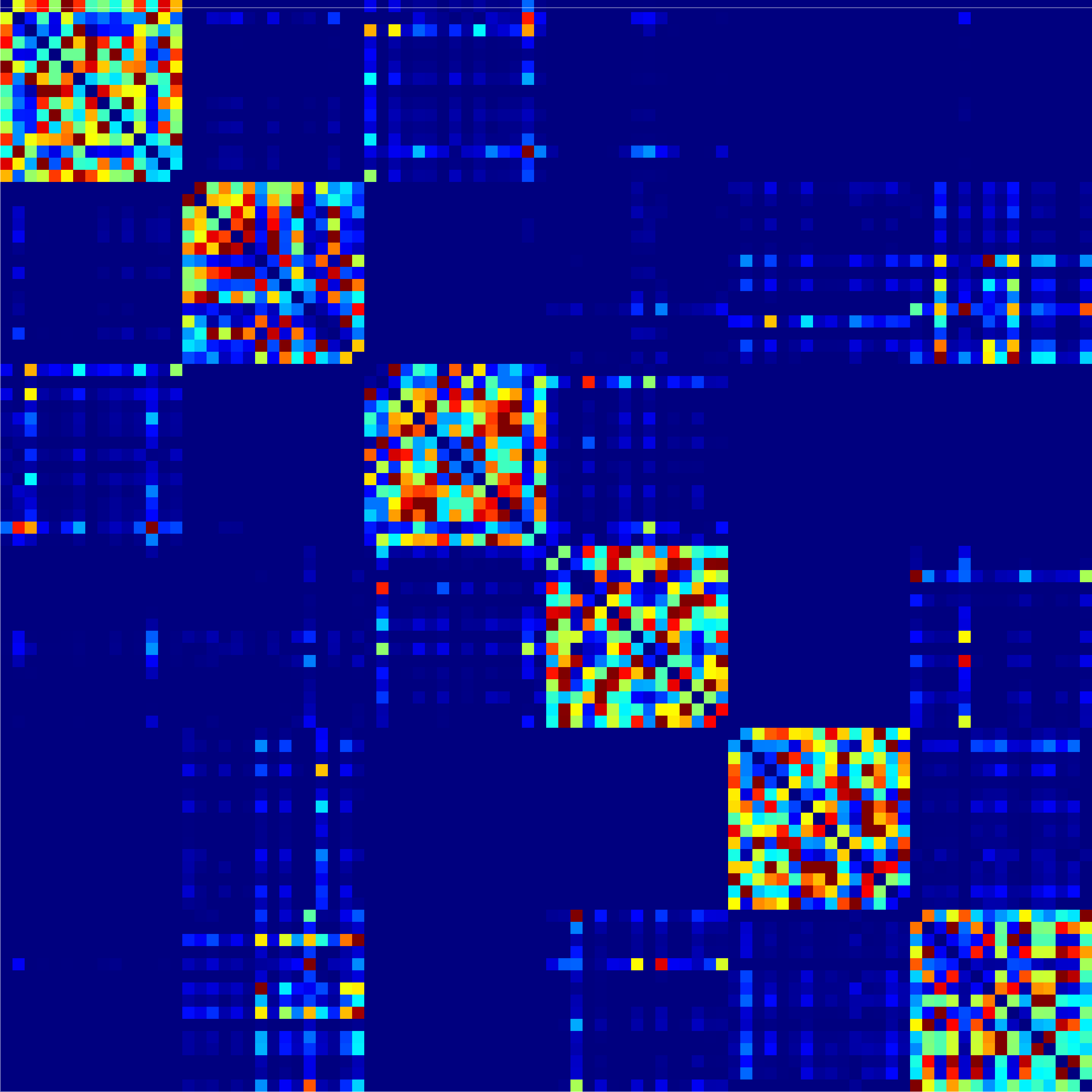}%{LSLDkernel_mutli_cluster_per_class}
        \caption{ }
        \label{fig_mulit_LSLD}
    \end{subfigure}
    \hfill
    \begin{subfigure}{0.3\columnwidth}
        \centering
        \includegraphics[width=\linewidth]{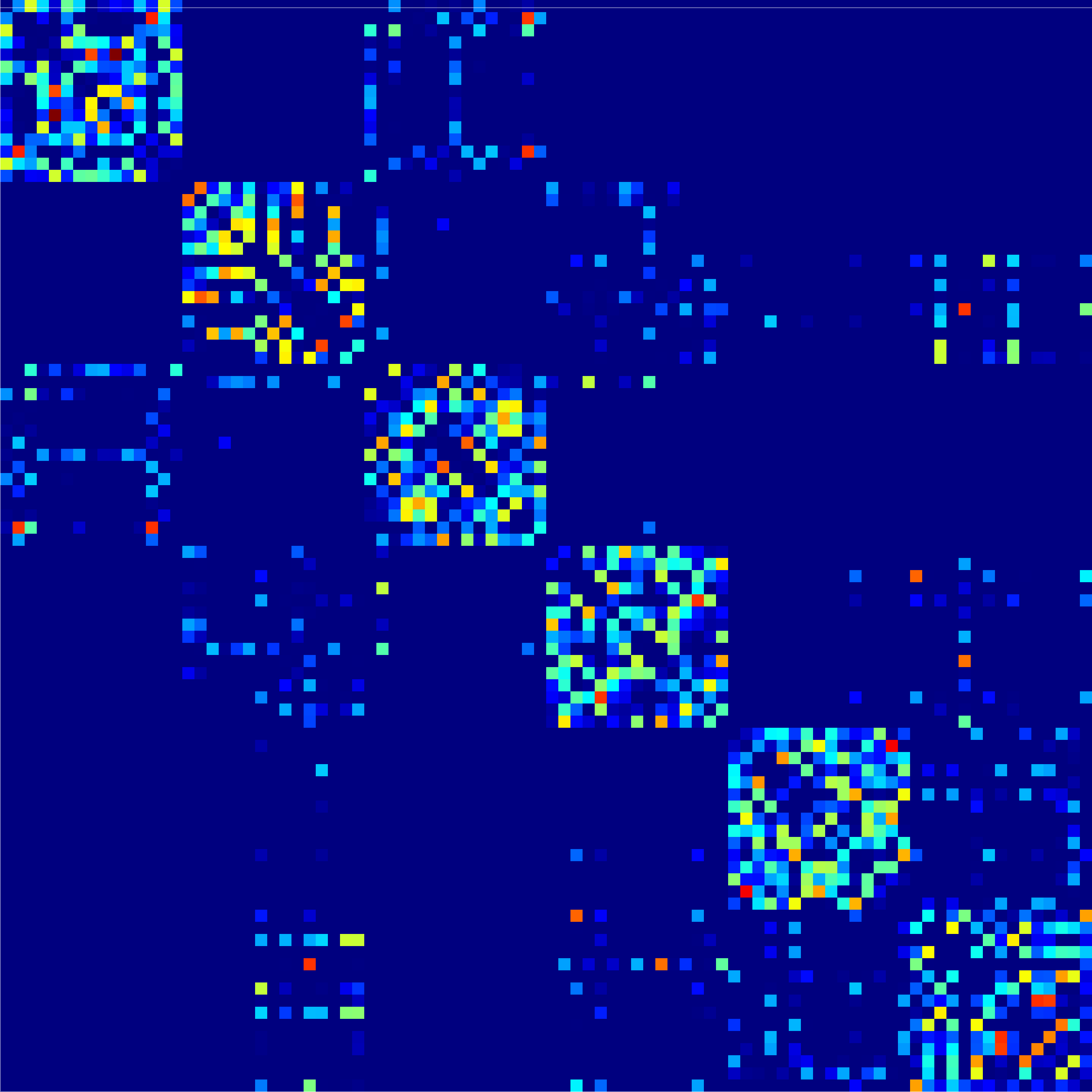}%{learned_ideal_kernel}
        \caption{ }
        \label{fig_learned_ideal}
    \end{subfigure}
	 \vspace{0.5em}    
    
    \begin{subfigure}{0.3\columnwidth}
        \centering
        \includegraphics[width=\linewidth]{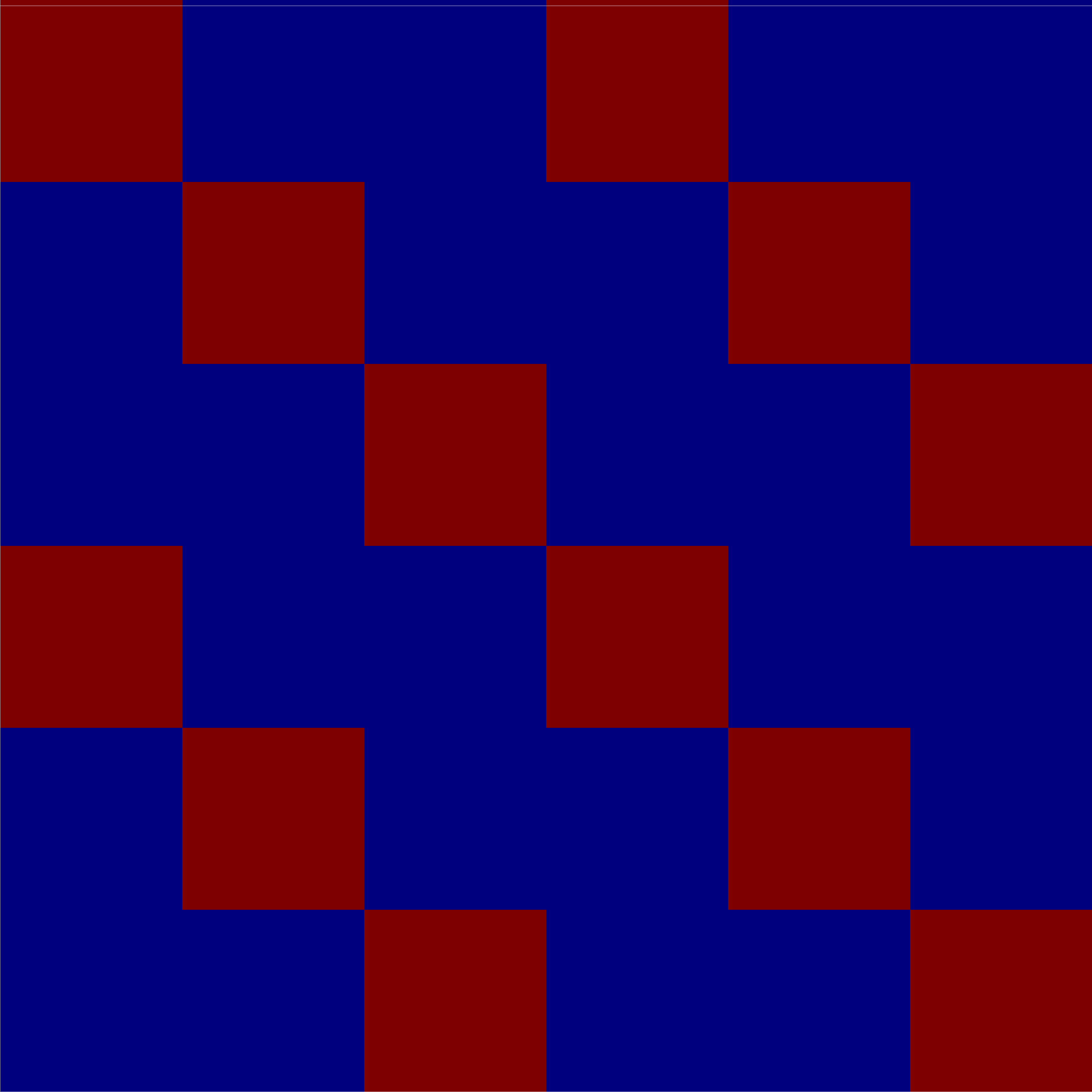}%{ideal_kernel_half_moons}
        \caption{ }
        \label{fig_ideal_half_moons}
    \end{subfigure}
    \hfill
    \begin{subfigure}{0.3\columnwidth}
        \centering
        \includegraphics[width=\linewidth]{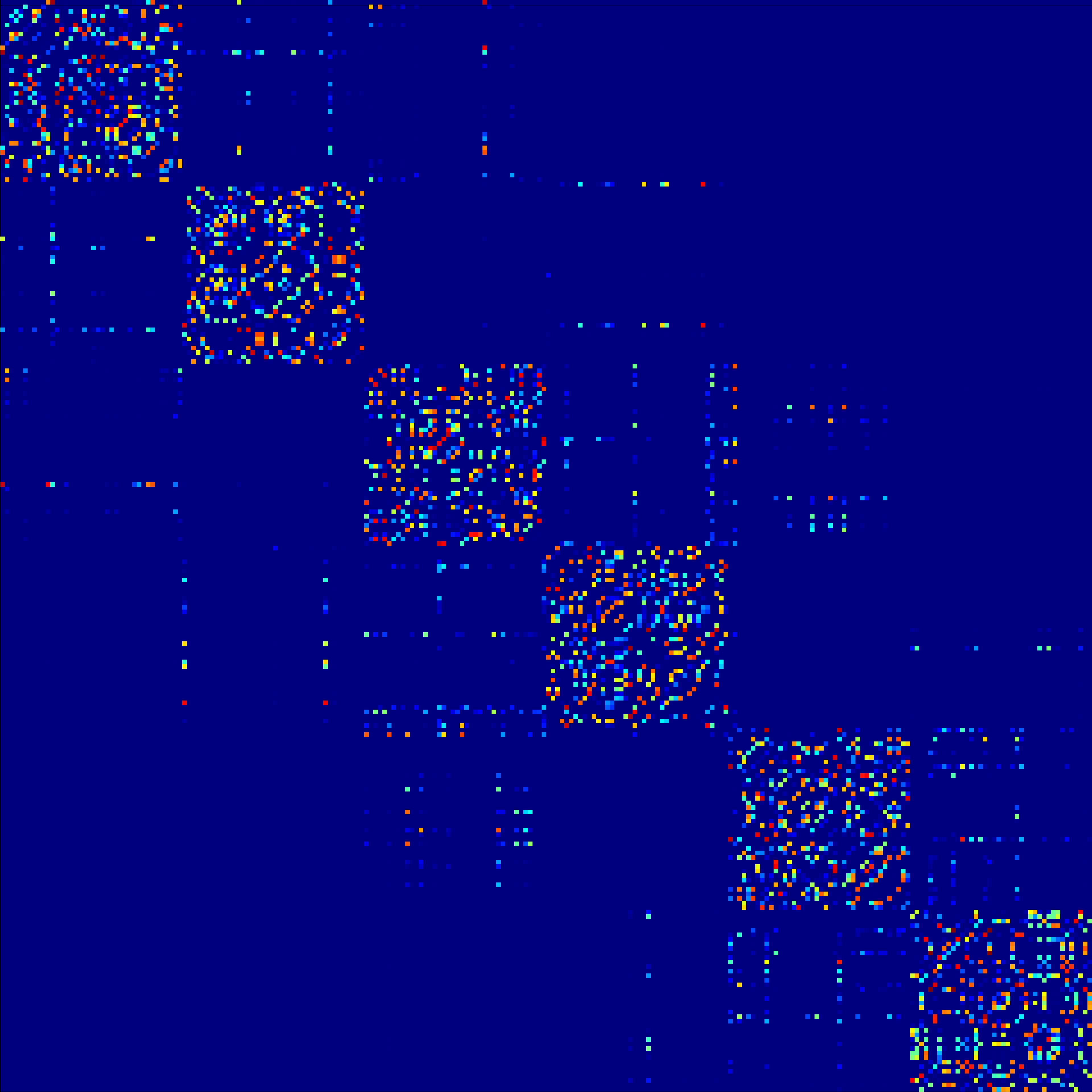}%{LSLD_half_moons}
        \caption{ }
        \label{fig_LSLD_half_moons}
    \end{subfigure}
    \hfill
    \begin{subfigure}{0.3\columnwidth}
        \centering
        \includegraphics[width=\linewidth]{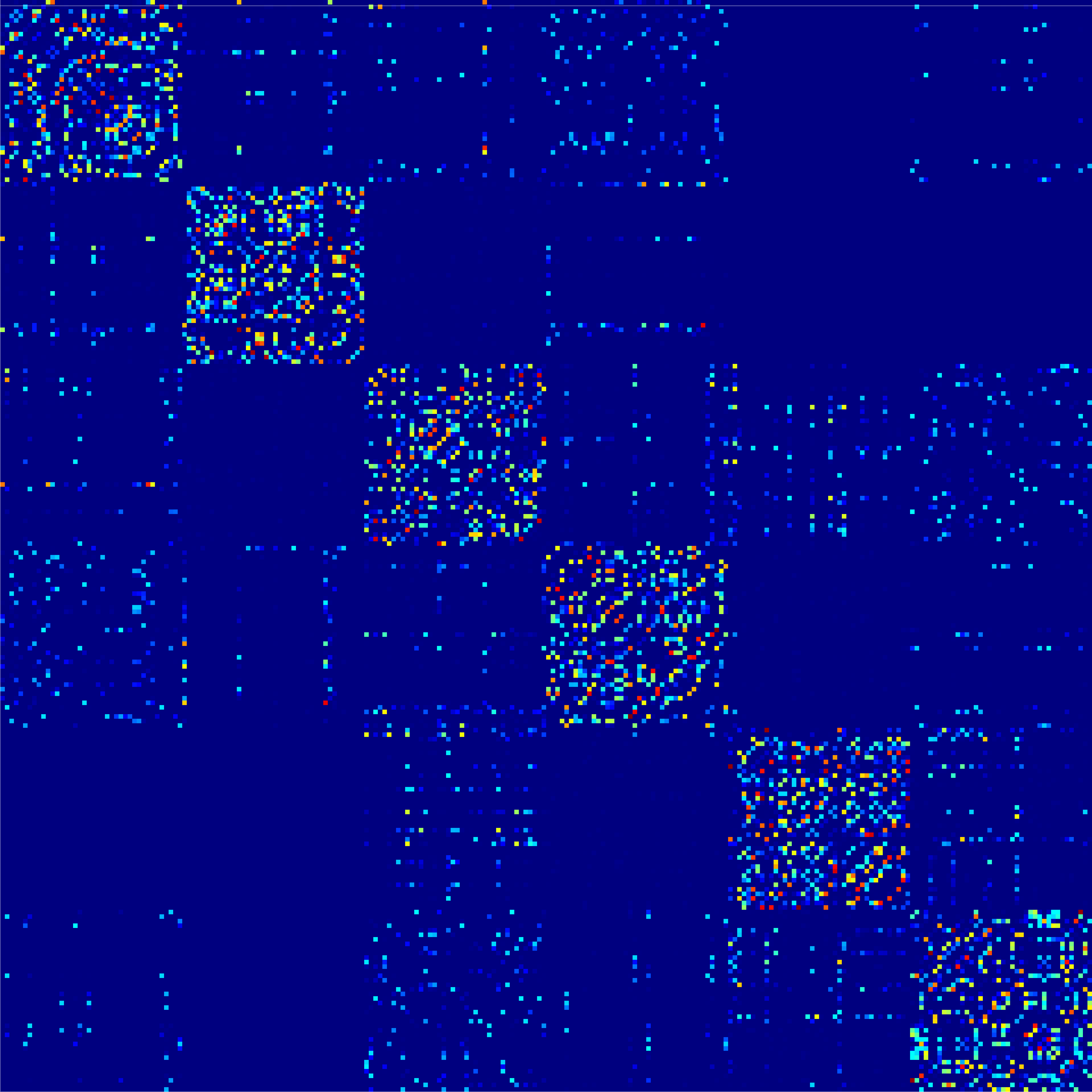}%{learned_ideal_halfmoons}
        \caption{ }
        \label{fig_learned_ideal_half_moons}
    \end{subfigure}
    
    \caption{Heat maps of kernels on the `Clusters' dataset (top) and `Noisy Crescent Moons' dataset (bottom). The ideal kernels are on the left, the LVBL kernels in centre, and the Learned to Align with the Ideal (LAI) kernels \cite{lanckriet2004learning} are on the right. Blue indicates minimum similarity and red indicates maximum similarity.}
    \label{fig_kernels}
\end{figure}

This section examines the relationship between classification accuracy and kernel measures across different kernel types, evaluating their ability to serve as absolute criteria for kernel selection. 

The two synthetic datasets presented in Fig.~\ref{fig_datasets} were designed with multimodal class structures to illustrate how the Confidence Score captures such structure.
Fig.~\ref{fig_multi_clust_per_class} presents the `Clusters' dataset; each class consists of three Gaussian clusters, and Fig.~\ref{fig_noisy_half_moons} presents the `Noisy Crescent Moons' dataset, where the data is distributed in intertwined crescent shapes with added noise. 
Fig.~\ref{fig_multi_ideal} and~\ref{fig_ideal_half_moons} present heat maps of the ideal kernels constructed for the `Clusters' and `Noisy Crescent Moons' datasets, respectively. 
Figures~\ref{fig_mulit_LSLD} and~\ref{fig_LSLD_half_moons} present a heat map of the LVBL kernels (see Equation \ref{eq_LSLDS}) for these datasets. 
Figures~\ref{fig_learned_ideal} and~\ref{fig_learned_ideal_half_moons} present the kernels learned by maximizing the Centred Kernel Alignment following \cite{lanckriet2004learning}; we refer to this as the Learn to Align with the Ideal (LAI) kernel.

%Note that the LAI kernel was constructed of a set that \forceemph{includes} the LVBL kernel, and so the LAI could be interpreted as an improvement on the LVBL kernel. %but did this additional stage of learning improve classification accuracy, and can kernel measures predict this improvement?
Table~\ref{tab_results_synthetic} compares kernel measures (CS, CKA, LKA, HKA\footnote{Here $\lambda = 0.5$ and the neighbourhood size is based on the fuzzy neighbourhoods captured by the kernel instead of a fixed value for $k$.}) and leave-one-out classification accuracy for the LVBL and LAI kernels on the `Clusters' and `Noisy Crescent Moon' datasets. 
On the `Clusters' dataset, the LAI kernel has a higher alignment to the ideal (CKA $= 0.283$, LKA $= 0.543$, HKA $= 0.413$) than the LVBL kernel (CKA $= 0.190$, LKA $= 0.426$, HKA $= 0.308$). 
This ordering is preserved no matter the value of $\lambda$. 
However, when comparing the Confidence Score, the LAI kernel has a lower Confidence Score (CS $= 0.898$) than the LVBL kernel (CS $= 0.937$).
As the LVBL kernel has the higher leave-one-out accuracy ($0.967$) than the LAI kernel ($0.922$), the CS is a better predictor of accuracy for these kernels on the `Clusters' dataset. 
The same pattern is evident for the `Noisy Crescent Moons' dataset.
%We do not use these results to prove that CS is always a better predictor of kernel classification accuracy, but to provided  2-dimensional illustrated examples of cases where the CS does better predict kernel classification accuracy over alignment to the ideal based measures.

\begin{table}[htbp]
\centering
\caption{Ideal kernel alignment-based and confidence based kernel measures on kernels, classification accuracy, on synthetic datasets from Fig.~\ref{fig_datasets}.}
\label{tab_results_synthetic}
\begin{tabular}{lccccc}
\toprule
  & CS & CKA & LKA & HKA & Acc. \\
\midrule
\makecell[l]{Cluster LVBL} & $\best{0.937}$ & $0.190$ & $0.426$ & $0.308$ & $\best{0.978}$ \\
\makecell[l]{Clusters LAI}  & $0.898$ & $\best{0.283}$ & $\best{0.543}$ & $\best{0.413}$ & $0.933$ \\
\midrule
\makecell[l]{Moons LVBL} & $\best{0.867}$ & $0.158$ & $0.378$ & $0.268$ & $\best{0.961}$ \\
\makecell[l]{Moons LAI}  & $0.821$ & $\best{0.259}$ & $\best{0.487}$ & $\best{0.373}$ & $0.931$ \\
%Noisy Half Moons LVBL & 0.170 & 0.769 & 0.188 & 0.274 & 0.209 & 0.948 \\
%Noisy Half Moons Learned Ideal & 0.202 & 0.719 & 0.420 & 0.491  & 0.456 & 0.926 \\
\bottomrule
\end{tabular}
\end{table}

\section{Future Work}
\label{sec_future_work}
%Fig.~\ref{fig_iris_many_kernel_measures} presents an evaluation of the bandwidth-tuning capabilities of these variants.
%Consistent with the results in Section~\ref{sec_tuning_bandwidth}, the Confidence Score identifies the bandwidth that produces the kernel with the highest classification accuracy. 
%The maximum of $\CS_A$ identifies the next-best bandwidth.
%For comparison, Fig.~\ref{fig_iris_many_kernel_measures} also includes CKA and a variant of CKA where the kernels are normalized. %in which the kernels are normalized. %using the procedure described in Section~\ref{sec_confidence} before computing CKA as defined in Equation~\ref{eq_cka}.
%The maximum of these kernel measures identify the third-best bandwidth.
%Finally, $\CS_B$ identifies the fourth-best bandwidth.
%The location of the maximum kernel measure is affected by the method of normalization. 
%Before drawing conclusions about the efficacy of $\CS_A$ and $\CS_B$, their theoretical properties should be further explored, and additional experiments on benchmark datasets across a range of tasks should be conducted.
 %, where the CS may serve as an  objective function for learning kernel combinations which improve classification accuracy.
Centered Kernel Alignment has found applicability in evaluating kernels induced by neural networks\cite{kornblith2019similarity, canatar2022kernel, liu2026humanguidenet}. Future work may investigate the use of CS as a measure of how well neural network representations capture the underlying class structure at different stages of the network. 
Additional future work will investigate the use of the CS in an objective function for Multiple Kernel Learning. 
%While the use of CS was proven to work well in kernel parameter tuning and kernel selection, additional future work would involve the evaluation of its usefulness in task such as in the evaluations of neural networks and feature selection/extraction.
%Evaluations of kernels constructed by neural networks in particular is a popular topic that we hope the Confidence Score will be relevant for.}
A further direction for future research is to extend the theoretical analysis by following the approach of \cite{cortes2009learning} to derive tighter error bounds. 
%Under suitable assumptions on the underlying data distribution, it may be possible to derive approximation guarantees or error bounds relating the Confidence Score to the true class structure. 

\section{Conclusion} 
\label{sec_conclusion}
%The Confidence Score showed superior performance to the Centred Kernel Alignment in kernel parameter learning and MKL applications. 
In Section~\ref{sec_eval}, we compared the performance of the Confidence Score to other kernel measures, finding in many cases that it was a better predictor of kernel classification accuracy. 
%The Confidence Score shows a significantly stronger correlation to the classification accuracy than the CKA. 
When used for kernel bandwidth tuning, the Confidence Score consistently identified bandwidths that produced kernels with higher classification accuracy than the bandwidths identified by Centred Kernel Alignment. 
Similarly, when used for kernel selection, the Confidence Score consistently identified kernels with higher classification accuracy and exhibited higher correlation to classification accuracy than kernel alignment-based measures and the regularized Fisher Separability Criterion. 
These results suggest the `similar $\rightarrow$ same class' used by the Confidence Score assumption allows for the identification of more accurate kernels than the `similar $\leftrightarrow$ same class' assumption or the `single-cluster-per-class' assumption used by other kernel measures.

Although the proposed Confidence Score demonstrated strong agreement with classification performance across the evaluated problems, the conclusions of this study are necessarily limited by the kernel families and datasets considered. 
\appendix

\section{Algorithm}
\label{sec_algorithm}
Algorithm~\ref{alg_fast_cs} computes the Confidence Score directly without explicitly constructing the hollowed matrix. 
\begin{algorithm}[htbp]
\caption{Fast computation of the Confidence Score}
\label{alg_fast_cs}

\begin{algorithmic}[1]
\Require Normalized kernel matrix $\ve{K}\in\mathbb{R}^{J\times J}$; class matrix $\ve{G}\in\mathbb{R}^{J\times C}$
\Ensure Confidence Score

\State $\mathrm{innerprod} \gets 0$
\State $\mathrm{norm} \gets 0$

\For{$i \gets 1$ to $J$}
    \For{$j \gets 1$ to $C$}
        \State $s \gets 0$
        \For{$k \gets 1$ to $J$}
            \If{$i \neq k$}
                \State $s \gets s + K_{ik}G_{kj}$
            \EndIf
        \EndFor
        \State $\mathrm{norm} \gets \mathrm{norm} + s^2$
        \State $\mathrm{innerprod} \gets \mathrm{innerprod} + sG_{ij}$
    \EndFor
\EndFor

\If{$\mathrm{innerprod} \leq \varepsilon$}
    \State \Return $0$
\EndIf

\State \Return $\displaystyle\frac{\mathrm{innerprod}}
{\sqrt{J}\sqrt{\mathrm{norm}}}$
\end{algorithmic}
\end{algorithm}


\begin{thebibliography}{99}

\bibitem{zhang2007local}
J.~Zhang, M.~Marsza{\l}ek, S.~Lazebnik, and C.~Schmid, ``Local features and
  kernels for classification of texture and object categories: A comprehensive
  study,'' \emph{International journal of computer vision}, vol.~73, pp.
  213--238, 2007.

\bibitem{gillespie2001optimizing}
B.~W. Gillespie and L.~E. Atlas, ``Optimizing time-frequency kernels for
  classification,'' \emph{IEEE Transactions on Signal Processing}, vol.~49,
  no.~3, pp. 485--496, 2001.

\bibitem{cortes2010two}
C.~Cortes, M.~Mohri, and A.~Rostamizadeh, ``Two-stage learning kernel
  algorithms.'' in \emph{ICML}, 2010, pp. 239--246.

\bibitem{hofmann2008kernel}
T.~Hofmann, B.~Sch{\"o}lkopf, and A.~J. Smola, ``Kernel methods in machine
  learning,'' 2008.

\bibitem{du2024exploring}
K.-L. Du, B.~Jiang, J.~Lu, J.~Hua, and M.~Swamy, ``Exploring kernel machines
  and support vector machines: Principles, techniques, and future directions,''
  \emph{Mathematics}, vol.~12, no.~24, p. 3935, 2024.

\bibitem{wang2017multiple}
Y.~Wang, X.~Liu, Y.~Dou, Q.~Lv, and Y.~Lu, ``Multiple kernel learning with
  hybrid kernel alignment maximization,'' \emph{Pattern Recognition}, vol.~70,
  pp. 104--111, 2017.

\bibitem{kumar2012binary}
A.~Kumar, A.~Niculescu-Mizil, K.~Kavukcuoglu, and H.~Daume~III, ``A binary
  classification framework for two-stage multiple kernel learning,''
  \emph{arXiv preprint arXiv:1206.6428}, 2012.

\bibitem{zhong2013optimizing}
S.~Zhong, D.~Chen, Q.~Xu, and T.~Chen, ``Optimizing the gaussian kernel
  function with the formulated kernel target alignment criterion for two-class
  pattern classification,'' \emph{Pattern Recognition}, vol.~46, no.~7, pp.
  2045--2054, 2013.

\bibitem{keerthi2003asymptotic}
S.~S. Keerthi and C.-J. Lin, ``Asymptotic behaviors of support vector machines
  with gaussian kernel,'' \emph{Neural computation}, vol.~15, no.~7, pp.
  1667--1689, 2003.

\bibitem{camargo2009multi}
J.~E. Camargo and F.~A. Gonz{\'a}lez, ``A multi-class kernel alignment method
  for image collection summarization,'' in \emph{Iberoamerican Congress on
  Pattern Recognition}.\hskip 1em plus 0.5em minus 0.4em\relax Springer, 2009,
  pp. 545--552.

\bibitem{tian2015efficient}
M.~Tian and W.~Wang, ``An efficient gaussian kernel optimization based on
  centered kernel polarization criterion,'' \emph{Information Sciences}, vol.
  322, pp. 133--149, 2015.

\bibitem{cristianini2001kernel}
N.~Cristianini, J.~Shawe-Taylor, A.~Elisseeff, and J.~Kandola, ``On
  kernel-target alignment,'' \emph{Advances in neural information processing
  systems}, vol.~14, 2001.

\bibitem{kandola2002optimizing}
J.~Kandola, J.~Shawe-Taylor, and N.~Cristianini, ``Optimizing kernel alignment
  over combinations of kernel,'' 2002.

\bibitem{lanckriet2004learning}
G.~R. Lanckriet, N.~Cristianini, P.~Bartlett, L.~E. Ghaoui, and M.~I. Jordan,
  ``Learning the kernel matrix with semidefinite programming,'' \emph{Journal
  of Machine learning research}, vol.~5, no. Jan, pp. 27--72, 2004.

\bibitem{lu2014multiple}
Y.~Lu, L.~Wang, J.~Lu, J.~Yang, and C.~Shen, ``Multiple kernel clustering based
  on centered kernel alignment,'' \emph{Pattern Recognition}, vol.~47, no.~11,
  pp. 3656--3664, 2014.

\bibitem{wang2018two}
T.~Wang, J.~Lu, and G.~Zhang, ``Two-stage fuzzy multiple kernel learning based
  on hilbert--schmidt independence criterion,'' \emph{IEEE Transactions on
  Fuzzy Systems}, vol.~26, no.~6, pp. 3703--3714, 2018.

\bibitem{kwok2003learning}
J.~T. Kwok and I.~W. Tsang, ``Learning with idealized kernels,'' in
  \emph{Proceedings of the 20th International Conference on Machine Learning
  (ICML-03)}, 2003, pp. 400--407.

\bibitem{cortes2009learning}
C.~Cortes, M.~Mohri, and A.~Rostamizadeh, ``Learning non-linear combinations of
  kernels,'' \emph{Advances in neural information processing systems}, vol.~22,
  2009.

\bibitem{mercer1909xvi}
J.~Mercer, ``Xvi. functions of positive and negative type, and their connection
  the theory of integral equations,'' \emph{Philosophical transactions of the
  royal society of London. Series A, containing papers of a mathematical or
  physical character}, vol. 209, no. 441-458, pp. 415--446, 1909.

\bibitem{suthaharan2016support}
S.~Suthaharan,
``Support vector machine,''
in \emph{Machine Learning Models and Algorithms for Big Data Classification: Thinking with Examples for Effective Learning},
pp.~207--235,
Springer, 2016.

\bibitem{scholkopf2001generalized}
B.~Sch{\"o}lkopf, R.~Herbrich, and A.~J. Smola, ``A generalized representer
  theorem,'' in \emph{International conference on computational learning
  theory}.\hskip 1em plus 0.5em minus 0.4em\relax Springer, 2001, pp. 416--426.

\bibitem{wang2015overview}
T.~Wang, D.~Zhao, and S.~Tian, ``An overview of kernel alignment and its
  applications,'' \emph{Artificial Intelligence Review}, vol.~43, pp. 179--192,
  2015.

\bibitem{ramona2012multiclass}
M.~Ramona, G.~Richard, and B.~David, ``Multiclass feature selection with kernel
  gram-matrix-based criteria,'' \emph{IEEE transactions on neural networks and
  learning systems}, vol.~23, no.~10, pp. 1611--1623, 2012.

\bibitem{islam2025kernel}
M.~T. Islam, D.~Liu, and D.~Sarkar, ``Kernel alignment using manifold
  approximation,'' in \emph{Second Workshop on Representational Alignment at
  ICLR 2025}.

\bibitem{cardenas2016centered}
D.~C{\'a}rdenas-Pe{\~n}a, D.~Collazos-Huertas, and G.~Castellanos-Dominguez,
  ``Centered kernel alignment enhancing neural network pretraining for
  mri-based dementia diagnosis,'' \emph{Computational and mathematical methods
  in medicine}, vol. 2016, no.~1, p. 9523849, 2016.

\bibitem{tian2023continuous}
S.~Tian, W.~Li, X.~Ning, H.~Ran, H.~Qin, and P.~Tiwari, ``Continuous transfer
  of neural network representational similarity for incremental learning,''
  \emph{Neurocomputing}, vol. 545, p. 126300, 2023.


\bibitem{alvarez2022gaussian}
S.~A. Alvarez, ``Gaussian rbf centered kernel alignment (cka) in the
  large-bandwidth limit,'' \emph{IEEE transactions on pattern analysis and
  machine intelligence}, vol.~45, no.~5, pp. 6587--6593, 2022.

\bibitem{kornblith2019similarity}
S.~Kornblith, M.~Norouzi, H.~Lee, and G.~Hinton, ``Similarity of neural network
  representations revisited,'' in \emph{International conference on machine
  learning}.\hskip 1em plus 0.5em minus 0.4em\relax PMlR, 2019, pp. 3519--3529.

\bibitem{canatar2022kernel}
A.~Canatar and C.~Pehlevan,
``A kernel analysis of feature learning in deep neural networks,''
in \emph{Proc. 58th Annual Allerton Conf. on Communication, Control, and Computing (Allerton)},
IEEE, pp.~1--8, 2022.

\bibitem{kang2026spectral}
H.~Kang, A.~Canatar, and S.~Chung,
``Spectral analysis of representational similarity with limited neurons,''
\emph{Advances in Neural Information Processing Systems},
vol.~38, pp.~24349--24383, 2026.

\bibitem{liu2026humanguidenet}
X.~Liu, Y.~Yang, and Z.~Zhang,
``HumanGuideNet: Adapter-Based Alignment of Deep Neural Networks with Human Similarity Judgments,''
in \emph{Proc. IEEE/CVF Winter Conf. on Applications of Computer Vision (WACV)},
pp.~1798--1808, 2026.
\bibitem{vance2024refining}
N.~Vance and P.~Flynn, ``Refining remote photoplethysmography architectures
  using cka and empirical methods,'' in \emph{Proceedings of the IEEE/CVF
  Conference on Computer Vision and Pattern Recognition}, 2024, pp. 380--388.

\bibitem{hu2024mpruner}
S.~Hu, C.~Park, A.~Ferraiuolo, S.-K. Ko, J.~Kim, H.~Song, and J.~Kim,
  ``Mpruner: Optimizing neural network size with cka-based mutual information
  pruning,'' \emph{arXiv preprint arXiv:2408.13482}, 2024.

\bibitem{wong2009constructive}
W.~W. Wong and F.~J. Burkowski, ``A constructive approach for discovering new
  drug leads: Using a kernel methodology for the inverse-qsar problem,''
  \emph{Journal of cheminformatics}, vol.~1, no.~1, p.~4, 2009.

\bibitem{kriegeskorte2008representational}
N.~Kriegeskorte, M.~Mur, and P.~A. Bandettini, ``Representational similarity
  analysis-connecting the branches of systems neuroscience,'' \emph{Frontiers
  in systems neuroscience}, vol.~2, p. 249, 2008.

\bibitem{williams2024equivalence}
A.~H. Williams, ``Equivalence between representational similarity analysis,
  centered kernel alignment, and canonical correlations analysis,''
  \emph{bioRxiv}, pp. 2024--10, 2024.
  
  \bibitem{ding2019identification}
Y.~Ding, J.~Tang, and F.~Guo, ``Identification of drug-side effect association
  via multiple information integration with centered kernel alignment,''
  \emph{Neurocomputing}, vol. 325, pp. 211--224, 2019.

\bibitem{raghu2017svcca}
M.~Raghu, J.~Gilmer, J.~Yosinski, and J.~Sohl-Dickstein, ``Svcca: Singular
  vector canonical correlation analysis for deep learning dynamics and
  interpretability,'' \emph{Advances in neural information processing systems},
  vol.~30, 2017.

\bibitem{morcos2018insights}
A.~Morcos, M.~Raghu, and S.~Bengio, ``Insights on representational similarity
  in neural networks with canonical correlation,'' \emph{Advances in neural
  information processing systems}, vol.~31, 2018.

\bibitem{nguyen2007kernel}
C.~H. Nguyen and T.~B. Ho, ``Kernel matrix evaluation.'' in \emph{IJCAI}, 2007,
  pp. 987--992.

\bibitem{xiong2005optimizing}
H.~Xiong, M.~Swamy, and M.~O. Ahmad, ``Optimizing the kernel in the empirical
  feature space,'' \emph{IEEE transactions on neural networks}, vol.~16, no.~2,
  pp. 460--474, 2005.
    
\bibitem{kim2006optimal}
S.-J. Kim, A. Magnani, and S. Boyd, ``Optimal kernel selection in kernel Fisher discriminant analysis,'' in \emph{Proc. 23rd Int. Conf. Mach. Learn. (ICML)}, 2006, pp. 465--472.

  \bibitem{li2003efficient}
H.~Li, T.~Jiang, and K.~Zhang,
``Efficient and robust feature extraction by maximum margin criterion,''
\emph{Advances in Neural Information Processing Systems},
vol.~16, 2003.

\bibitem{wan2023global}
H.~Wan, H.~Wang, B.~W. Scotney, J.~Liu, and X.~Wei, ``Global subclass
  discriminant analysis,'' \emph{Knowledge-Based Systems}, vol. 280, p. 111010,
  2023.

\bibitem{wan2020within}
H.~Wan, H.~Wang, B.~Scotney, J.~Liu, and W.~W. Ng, ``Within-class multimodal
  classification,'' \emph{Multimedia Tools and Applications}, vol.~79, no.~39,
  pp. 29\,327--29\,352, 2020.

\bibitem{hofer2022model}
E.~Hofer and M.~v.~Mohrenschildt, ``Model-free data mining of families of
  rotating machinery,'' \emph{Applied Sciences}, vol.~12, no.~6, p. 3178, 2022.

\bibitem{reynolds2000speaker}
D.~A. Reynolds, T.~F. Quatieri, and R.~B. Dunn, ``Speaker verification using
  adapted gaussian mixture models,'' \emph{Digital signal processing}, vol.~10,
  no. 1-3, pp. 19--41, 2000.

\bibitem{blekas2005spatially}
K.~Blekas, A.~Likas, N.~P. Galatsanos, and I.~E. Lagaris, ``A spatially
  constrained mixture model for image segmentation,'' \emph{IEEE transactions
  on Neural Networks}, vol.~16, no.~2, pp. 494--498, 2005.

\bibitem{chung1997spectral}
F.~R. Chung, \emph{Spectral graph theory}.\hskip 1em plus 0.5em minus
  0.4em\relax American Mathematical Soc., 1997, vol.~92.

\bibitem{belkin2005manifold}
M.~Belkin, P.~Niyogi, and V.~Sindhwani, ``On manifold regularization,'' in
  \emph{International Workshop on Artificial Intelligence and
  Statistics}.\hskip 1em plus 0.5em minus 0.4em\relax PMLR, 2005, pp. 17--24.
  
  \bibitem{cortes2013learning}
C.~Cortes, M.~Kloft, and M.~Mohri,
``Learning kernels using local Rademacher complexity,''
\emph{Advances in Neural Information Processing Systems},
vol.~26, 2013.

\bibitem{koltchinskii2002empirical}
V.~Koltchinskii and D.~Panchenko, ``Empirical margin distributions and bounding the generalization error of combined classifiers,'' \emph{The Annals of Statistics}, vol.~30, no.~1, pp.~1--50, 2002.

\bibitem{bartlett2002rademacher}
P.~L. Bartlett and S.~Mendelson, ``Rademacher and Gaussian complexities: Risk bounds and structural results,'' \emph{Journal of Machine Learning Research}, vol.~3, pp.~463--482, 2002.

\bibitem{mohri2018foundations}
M.~Mohri, A.~Rostamizadeh, and A.~Talwalkar, \emph{Foundations of Machine Learning}, 2nd~ed. Cambridge, MA, USA: MIT Press, 2018.

\bibitem{coifman2006diffusion}
R.~R. Coifman and S.~Lafon,
``Diffusion maps and coarse-graining: A unified framework for dimensionality reduction, graph partitioning, and data set parameterization,''
\emph{IEEE Transactions on Pattern Analysis and Machine Intelligence},
vol.~28, no.~4, pp.~569--582, 2006.

\bibitem{yousefi2018local}
N.~Yousefi, Y.~Lei, M.~Kloft, M.~Mollaghasemi, and G.~C. Anagnostopoulos, ``Local Rademacher complexity-based learning guarantees for multi-task learning,'' \emph{Journal of Machine Learning Research}, vol.~19, no.~38, pp.~1--47, 2018.

\bibitem{palar2023kernel}
P.~S. Palar, L.~Parussini, L.~Bregant, K.~Shimoyama, and L.~R. Zuhal, ``On
  kernel functions for bi-fidelity gaussian process regressions,''
  \emph{Structural and Multidisciplinary Optimization}, vol.~66, no.~2, p.~37,
  2023.
  
  \bibitem{wong2011using}
W. W. L. Wong and F. J. Burkowski,
``Using Kernel Alignment to Select Features of Molecular Descriptors in a QSAR Study,''
\emph{IEEE/ACM Transactions on Computational Biology and Bioinformatics},
vol.~8, no.~5, pp.~1373--1384, 2011.

\bibitem{muyskens2024identifiability}
A.~Muyskens, B.~W. Priest, I.~R. Goumiri, and M.~D. Schneider,
  ``Identifiability and sensitivity analysis of kriging weights for the matern
  kernel,'' \emph{arXiv preprint arXiv:2410.08310}, 2024.

\bibitem{lu2025choice}
J.~Lu, Q.~J. Wang, N.~Fraehr, X.~Xiang, and X.~Wu, ``Choice of gaussian process
  kernels used in lsg models for flood inundation predictions,'' \emph{Journal
  of Hydrology}, vol. 655, p. 132949, 2025.
  

\bibitem{hofer2024locally}
E.~Hofer and M.~v.~Mohrenschildt, ``Locally-scaled kernels and confidence
  voting,'' \emph{Machine Learning and Knowledge Extraction}, vol.~6, no.~2,
  pp. 1126--1144, 2024.

\bibitem{mcinnes2018umap}
L.~McInnes, J.~Healy, and J.~Melville, ``Umap: Uniform manifold approximation
  and projection for dimension reduction,'' \emph{arXiv preprint
  arXiv:1802.03426}, 2018.

\bibitem{Dua:2017}
%\BIBentryALTinterwordspacing
D.~Dheeru and E.~Karra~Taniskidou, ``{UCI} machine learning repository,'' 2017.
  [Online]. Available: \url{http://archive.ics.uci.edu/ml}
%\BIBentrySTDinterwordspacing

\bibitem{deng2012mnist}
L.~Deng, ``The MNIST database of handwritten digit images for machine learning research [Best of the Web],'' \emph{IEEE Signal Processing Magazine}, vol.~29, no.~6, pp.~141--142, Nov.~2012.

\bibitem{chudzian2012evaluation}
P. Chudzian, ``Evaluation measures for kernel optimization,''
\emph{Pattern Recognition Letters}, vol. 33, no. 9,
pp. 1108--1116, 2012.

\bibitem{cocor}
B. Diedenhofen and J. Musch,
``cocor: A Comprehensive Solution for the Statistical Comparison of Correlations,''
\emph{PLoS ONE},
vol.~10, no.~4, p.~e0121945, 2015.
doi: 10.1371/journal.pone.0121945.

\bibitem{olsen2024think}
K.~Olsen, R.~M.~H. Lindrup, and M.~M{\o}rup,
``Think global, adapt local: Learning locally adaptive k-nearest neighbor kernel density estimators,''
in \emph{Proceedings of the International Conference on Artificial Intelligence and Statistics (AISTATS)},
pp.~4114--4122,
PMLR, 2024.

\end{thebibliography}
\end{document}